\documentclass[10pt,twocolumn,letterpaper]{article}

\usepackage{cvpr}              
\makeatletter
\@namedef{ver@everyshi.sty}{}
\makeatother
\usepackage{times}
\usepackage{epsfig}
\usepackage{graphicx}
\usepackage{amsmath}
\usepackage{amssymb}
\usepackage{bm}
\usepackage{multirow}
\usepackage{booktabs}
\usepackage{xcolor}
\usepackage{microtype}
\usepackage{enumitem}
\usepackage{algorithm}
\usepackage{mathtools}
\usepackage[noend]{algpseudocode}
\usepackage[accsupp]{axessibility}
\usepackage{tikz}
\usetikzlibrary{arrows,intersections}
\tikzset{font=\scriptsize}
\usepackage{pgfplots}
\pgfplotsset{compat=1.11}
\usepgfplotslibrary{fillbetween}
\usetikzlibrary{backgrounds}
\definecolor{plotsgreen}{RGB}{38,150,38}
\definecolor{plotsorange}{RGB}{255,115,17}
\definecolor{plotsred}{RGB}{208,34,35}
\definecolor{plotspurple}{RGB}{137,92,181}
\definecolor{plotsblue}{RGB}{0,101,167}

\usepackage{mathtools}

\DeclarePairedDelimiterX{\infdivx}[2]{(}{)}{%
  #1\;\delimsize\|\;#2%
}
\newcommand{\infdiv}{D_{KL}\infdivx}

\usepackage{array}
\newcolumntype{L}{>{\centering\arraybackslash}m{0.2\columnwidth}}
\newcolumntype{P}[1]{>{\centering\arraybackslash}p{#1}}
\newcommand{\red}{\textcolor{black}}

\newcommand{\YM}[1]{{\color{blue}[YM: #1]}}

\newcommand{\XY}[1]{{\color{magenta}[XY: #1]}}

\newcommand{\samchange}[1]{{\color{black} #1}}

\usepackage[pagebackref,breaklinks,colorlinks]{hyperref}
\usepackage{changes}

\usepackage[capitalize]{cleveref}
\crefname{section}{Sec.}{Secs.}
\Crefname{section}{Section}{Sections}
\Crefname{table}{Table}{Tables}
\crefname{table}{Tab.}{Tabs.}

\def\cvprPaperID{4852} 
\def\confName{CVPR}
\def\confYear{2022}

\begin{document}

\title{Controllable Dynamic Multi-Task Architectures}


\author{Dripta S. Raychaudhuri$^{1 }$ \quad Yumin Suh$^{2 }$ \quad Samuel Schulter$^{2 }$ \quad Xiang Yu$^{2 }$ \quad Masoud Faraki$^{2 }$ \\
Amit K. Roy-Chowdhury$^{1 }$ \quad Manmohan Chandraker$^{2, 3 }$\\
$^{1}$University of California, Riverside \quad \ $^{2}$NEC Labs America \quad \ $^{3}$University of California, San Diego
}

\maketitle

\begin{abstract}
\samchange{Multi-task learning commonly encounters competition for resources among tasks, specifically when model capacity is limited.  This challenge motivates models which allow control over the relative importance of tasks and total compute cost during inference time.}
In this work, we propose such a controllable multi-task network that dynamically adjusts its architecture and weights to match the desired task preference as well as the resource constraints. 
In contrast to the existing dynamic multi-task approaches that adjust only the weights within a fixed architecture,
%
our approach affords the flexibility to dynamically control the total computational cost and match the user-preferred task importance better.
\red{We propose a disentangled training of two hypernetworks, by exploiting task affinity and a novel branching regularized loss, to take input preferences and accordingly predict tree-structured models with adapted weights.}
Experiments on three multi-task benchmarks, namely PASCAL-Context, NYU-v2, and CIFAR-100, show the efficacy of our approach. Project page is available at \url{https://www.nec-labs.com/~mas/DYMU}.

\end{abstract}

\section{Introduction}
Multi-task learning~\cite{caruana1997multitask,ruder2017overview} (MTL) solves multiple tasks using a single model, with potential advantages of fast inference and improved generalization by sharing representations across related tasks.
\begin{figure}
    \centering
    \includegraphics[width=0.45\textwidth]{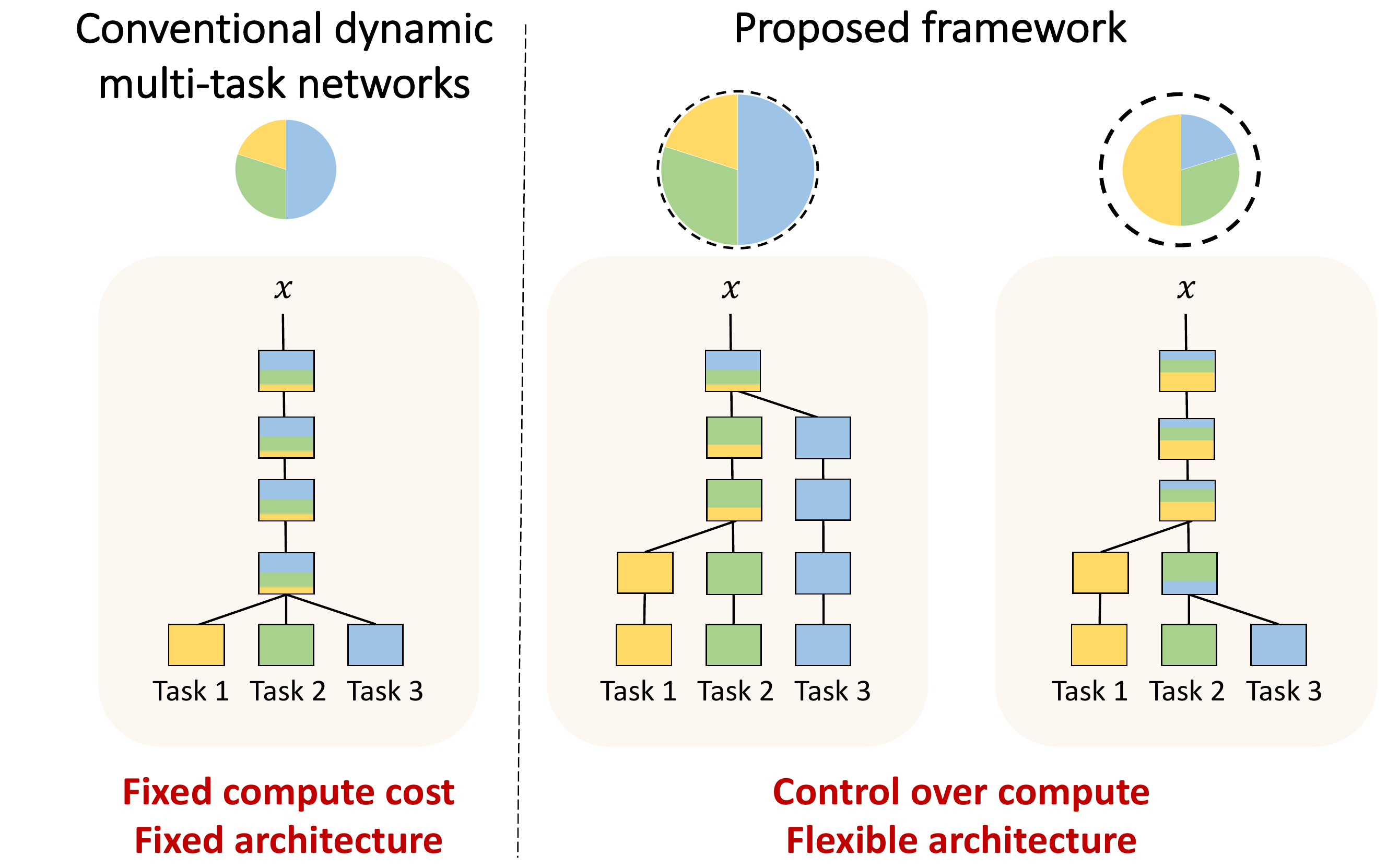}
    \vspace{-0.2cm}
    \caption{\textbf{Problem setup.} Our goal is to enable users to control resource allocation dynamically among multiple tasks at inference time. Conventional dynamic networks (PHN~\cite{navon2021learning}) for MTL achieve this in terms of weight changes within a fixed model (color gradients indicate proportion of weights allocated for each task). In contrast, we perform resource allocation in terms of both architecture and weights. This enables us to control total compute cost in addition to task preference. Dashed circle represents maximum compute budget, while filled circle represents the desired budget. Portion of colors represents the user-defined task importance.} 
    \vspace{-0.5cm}
    \label{fig:teaser}
\end{figure}
However, in practical scenarios, simultaneously optimizing all tasks is difficult due to task conflicts and limited model capacity~\cite{zamir2018taskonomy}. Consequently, a trade-off between the competing tasks has to be found, necessitating precise balancing of the different task losses during optimization. In many applications, the desired trade-off can change over time, requiring a new model to be retrained from scratch. To overcome this lack of flexibility, recent methods propose dynamic networks for multi-task learning~\cite{navon2021learning,lin2020controllable}. These frameworks enable a single multi-task model to learn the entire trade-off curve, and allow users to control the desired trade-off during inference via task preferences denoting the relative task importance.

Conventional dynamic approaches for MTL assume a fixed model architecture, with all but the last prediction layers shared, and control trade-offs by changing the weights of this model. While such hard-parameter sharing is helpful in saving resources, the performance is inevitably lower than single task baselines when task conflicts exist due to over-sharing of parameters between tasks~\cite{ruder2017overview} .
Furthermore, the fixed architecture suffers from a lack of flexibility, leading to a constant compute cost irrespective of the given task preference or compute budget changes. In many applications where the budget can change over time, these approaches may fail to take advantage of the increased resources in order to improve performance or accordingly lower the compute cost in order to satisfy stricter budget requirements.

To address the aforementioned issue and strike a balance between flexibility and performance, we propose a more expressive tree-structured~\cite{guo2020learning} dynamic multi-task network which can adapt its \emph{architecture} in addition to its weights at test-time, as illustrated in Figure~\ref{fig:teaser}.
Specifically, we design a controller using two hypernetworks~\cite{ha2016hypernetworks} that predict architectures and weights, respectively, given a user preference that specifies test-time trade-offs of relative {\em task importance} and {\em resource availability}.
This increases flexibility 
by changing branching locations to re-allocate resources over tasks to match user-preferred task importance, and enhance or compromise task accuracy given computation budget requirements at any given moment. 
However, this comes at the cost of increase in complexity:
1) generalizing architecture prediction to unseen preferences, and 2) performing dynamic weight changes on potentially thousands of different models.
To tackle these challenges, we develop a two-stage training scheme that starts from an $N$-stream network, termed the anchor net, which is initialized using weights from $N$ pre-trained single-task models.
This guides the architecture search as a prior that is preference-agnostic yet captures \emph{inter-task relations}.
In the first stage, we exploit inter-task relations derived from the anchor net to train the first hypernetwork that predicts connections between the different streams. We introduce \red{a branching regularized loss} that encourages more {\em resource allocation} for dominant tasks while reducing the network cost from the less preferred ones. The predicted architectures contain edges that have not been observed during the anchor net initialization. These are denoted as cross-task edges since they connect nodes that belong to different streams. In the second stage, to improve the performance of the predicted architectures with cross-task edges, we train a secondary hypernetwork for {\em cross-task adaptation} via modulation of the normalization parameters. 


Our framework is evaluated on three MTL datasets (PASCAL-Context, NYU-v2 and CIFAR-100) in terms of task performance, computational cost, and controllability (for both task importance and computational cost). 
Achieving performance comparable to state-of-the-art MTL architecture search methods under uniform task preference, our controller can further approximate efficient architectures for non-uniform preferences with provisions for reducing network size depending on computational constraints. 

The primary contributions of our work are as follows:

\begin{itemize}[leftmargin=*,noitemsep,nolistsep]
\item A controllable multi-task framework which allows users to assign task preference and the trade-off between task performance and network capacity via architectural changes. 
\item A controller, composed of two hypernetworks, to provide dynamic network structure and adapted network weights.
\item A new joint learning objective including task-related losses and network complexity regularization to achieve the user defined trade-offs.
\item Experiments on several MTL benchmarks (PASCAL-Context~\cite{mottaghi2014the},  NYU-v2~\cite{silberman2012indoor}, CIFAR-100~\cite{krizhevsky2009learning}) demonstrate the efficacy of our framework. 
\end{itemize}

\begin{figure*}[ht]
    \centering
    \includegraphics[width=0.9\textwidth]{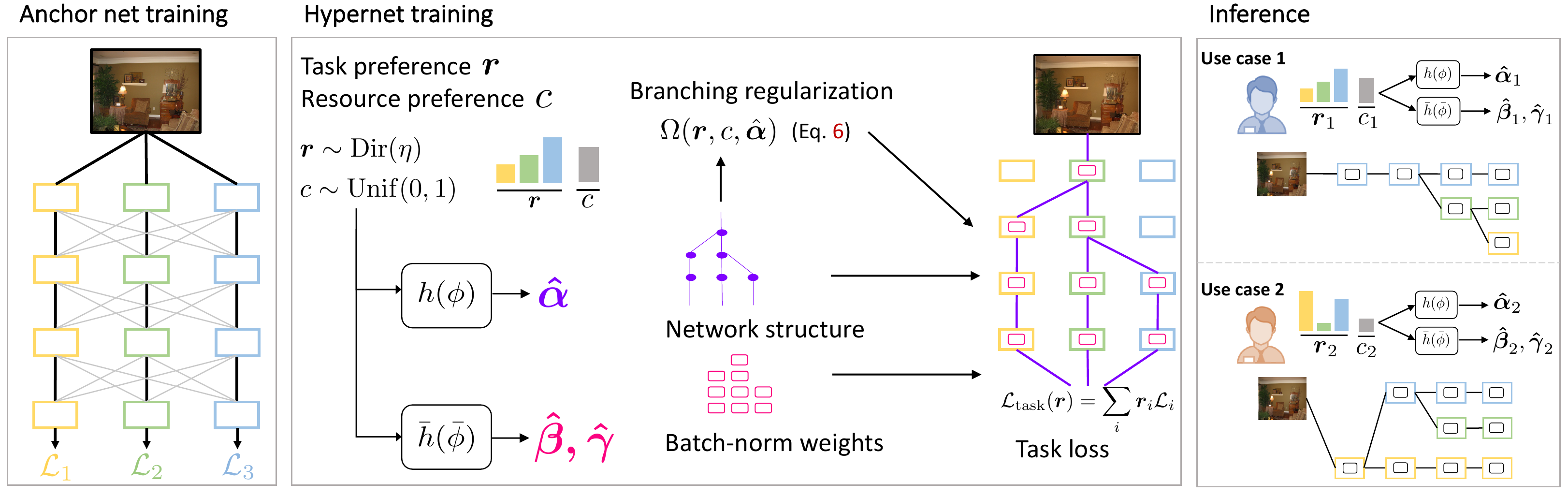}
    \vspace{-0.25cm}
    \caption{\textbf{Overview of framework.} We initialize our framework using an anchor net which consists of single-task networks. During training, we first train the edge hypernet $h(\phi)$ using sampled preferences $(\bm{r}, c)$ to optimize the task loss and a branching regularizer, for preference aware branching. Next, we optimize the weight hypernet $\bar{h}(\bar{\phi})$ in a similar fashion by minimizing only the task loss. At inference, the hypernets jointly predict architecture and weights according to the user preferences. 
    }
    \label{fig:framework}
    \vspace{-0.4cm}
\end{figure*}

\section{Related Work}
\noindent \textbf{Multi-Task Learning.} 
Multi-task learning seeks to learn a single model to simultaneously solve a variety of learning tasks by sharing information among the tasks \cite{caruana1997multitask}. In the context of deep learning, current works focus mostly on designing novel network architectures and constructing efficient shared representation among tasks \cite{ruder2017overview,zhang2021survey}. Typically, these works can be grouped into two classes - \emph{hard-parameter sharing} and \emph{soft-parameter sharing}. In the soft sharing setting \cite{misra2016cross,gao2019nddr,ruder2019latent}, each task has its own set of backbone parameters with some sort of regularization mechanisms to enforce the distance between weights of the model to be close. In contrast, the hard sharing setting entails all the tasks sharing the same set of backbone parameters, with branches towards the outputs \cite{long2015learning,kokkinos2017ubernet,kendall2018multi}. More recent works have attempted learning the optimal architectures via differentiable architecture search \cite{guo2020learning,bruggemann2020automated,sun2019adashare}. The overwhelming majority of these approaches are trained using a simple weighted sum of the individual task losses, where a proper set of weights is commonly selected using grid search or using techniques such as gradient balancing \cite{chen2018gradnorm}. 
Other approaches \cite{sener2018multi,lin2019pareto,mahapatra2020multi} attempt to model multi-task learning as a multi-objective optimization problem and find Pareto stationary solutions among different tasks. Recently, optimization methods have also been proposed to manipulate gradients in order to avoid conflicts across tasks~\cite{chen2020just,yu2020gradient}. None of these methods are suitable for dynamically modeling performance trade-offs, which is the focus of our work.

\noindent \textbf{Hypernetworks.}
A hypernetwork is used to learn context dependent parameters for a dynamic network \cite{ha2016hypernetworks,schmidhuber1992hypernet}, thus, obtaining multiple customizable models using a single network. Such hypernetworks have been successfully applied in different scenarios, \eg, recurrent networks \cite{ha2016hypernetworks}, 3D point cloud prediction \cite{littwin2019deep}, video frame prediction \cite{jia2016dynamic}, neural architecture search \cite{brock2017smash} and reinforcement learning \cite{rashid2018qmix,sarafian2021recomposing}. Recent works \cite{navon2021learning,lin2020controllable} propose using hypernetworks to model the Pareto front of competing multi-task objectives. Our approach is closely related to these works, however, these methods focus on generating weights for a fixed, handcrafted architecture, while we use hypernetworks to model the trade-offs in multi-task learning by varying the architecture. This allows us to take dynamic resource allocation into account, an aspect largely ignored in previous works.

\noindent \textbf{Dynamic Networks.}
Dynamic neural networks, as opposed to usual static models, can adapt their structures during inference, leading to notable improvements in performance and computational efficiency \cite{han2021dynamic}. Previous works focus on adjusting the network depth \cite{veit2018convolutional,bolukbasi2017adaptive,wang2018skipnet,huang2017multi}, width \cite{yuan2020s2dnas,li2021dynamic}, or perform dynamic routing within a fixed supernet that includes multiple possible paths \cite{liu2018dynamic,li2020learning,odena2017changing}. Dynamic depth is realised by either early exiting, \ie allowing ``easy'' samples to be processed at shallow layers without executing the deeper layers \cite{bolukbasi2017adaptive,huang2017multi}, or layer skipping, \ie selectively skipping intermediate network layers conditioned on each sample \cite{veit2018convolutional,wang2018skipnet}. Dynamic width is an alternative to the dynamic depth where instead of layers, filters are selectively pruned conditioned on the input  \cite{yuan2020s2dnas,li2021dynamic}. Dynamic routing can be implemented by learning controllers to selectively execute one of multiple candidate modules at each layer \cite{liu2018dynamic,odena2017changing}. Due to the non-differentiable nature of the discrete choices, reinforcement learning is employed to learn these controllers. In \cite{li2020learning}, the routing modules utilize a differentiable activation function which conditionally outputs zero values, facilitating the end-to-end training of routing decisions. Recent works have also proposed learning dynamic weights for modeling different hyperparameter configurations \cite{dosovitskiy2019you} and domain adaptation \cite{wang2020tent}. 
In contrast to most of the existing works which intrinsically adapt network structures as a function of input, our method enables explicit control of the total computational cost as well as the task trade-offs.

\noindent \textbf{Weight Sharing Neural Architecture Search.} Weight sharing has evolved as a powerful tool to amortize computational cost across models for neural architecture search (NAS). These methods integrate the whole search space of architectures into a weight sharing supernet and optimize network architectures by pursuing the best performing sub-networks. Joint optimization methods \cite{liu2018darts,wu2019fbnet,cai2018proxylessnas} optimize the weights of the supernet and a differentiable routing policy simultaneously. In contrast, one-shot methods \cite{bender2018understanding,brock2017smash,akimoto2019adaptive,guo2020single} disentangle the training into two steps: first, the weights of the supernet are trained, after which the agent is trained with the fixed supernet. We utilize such a weight sharing strategy in our framework for dynamic resource allocation.



\section{Method}
Given a set of $N$ tasks $\mathbf{T}=\{\mathcal{T}_1,\mathcal{T}_2,\dots,\mathcal{T}_N\}$, conventional multi-task learning seeks to minimize a weighted sum of task-specific losses: $\mathcal{L}_{\textrm{task}}(\bm{r}) = \sum_ir_i\mathcal{L}_i$, where each $\mathcal{L}_i$ represents the loss associated with task $\mathcal{T}_i$, and $\bm{r}$ denotes a task preference vector. This vector signifies the desired performance trade-off across the different tasks, with larger values of $r_i$ denoting higher importance to task $\mathcal{T}_i$. Here $\bm{r} \in \mathcal{S}_N$, where $\mathcal{S}_N=\{\bm{r}\in\mathbb{R}^N | \sum_i r_i = 1, r_i \geq 0\}$ represents the $N$-dimensional simplex~\cite{ruder2017overview}. 
We seek to approximate the trade-off curve defined by different values of $\bm{r}$ using tree-structured sub-networks~\cite{guo2020learning} within a single multi-task model, given a total computational budget defined by a resource preference variable $c \in [0,1]$, where larger $c$ denotes more frugal resource usage.
This is formulated as a minimization of the expected value of the task loss over the user preference distribution, with regularization $\Omega$ to control resource usage, \ie, $\mathbb{E}_{(\bm{r}, c)\sim P_{(\bm{r}, c)}}\mathcal{L}_{\textrm{task}}(\bm{r}) + \Omega(\bm{r}, c)$.
Optimizing this directly is equivalent to solving NAS~\cite{liu2018darts} for every possible $(\bm{r}, c)$ simultaneously.
Thus, instead of solving directly, we cast it as a search to find tree sub-structures and the corresponding modulation of features for every $(\bm{r}, c)$, within an $N$-stream anchor network with fixed weights.

Our framework consists of two \emph{hypernets} ($h$ and $\bar{h}$)~\cite{ha2016hypernetworks} and an \emph{anchor net} $\mathrm{F}$, as shown in Figure~\ref{fig:framework}.
At test-time, given an input preference, we utilize the network connections and adapted weights predicted by the hypernets to modulate $\mathrm{F}$, to obtain the final model.
We propose a two-stage training scheme to train the framework.
First, we initialize a preference agnostic anchor net, which provides the anchor weights at test time (Section~\ref{sec:supernet}).
Based on this anchor net, the tree-structured architecture search space is then defined (Section~\ref{sec:search_space}).
Next, we train the \emph{edge hypernet} using prior task relations obtained from the anchor net by optimizing a novel branching regularized loss function derived by inducing a dichotomy over the tasks (Section~\ref{sec:edge_hypernet}).
Finally, we train a \emph{weight hypernet}, keeping the anchor net and edge hypernet fixed, to modulate the anchor net weights (Section~\ref{sec:weight_hypernet}).

\subsection{Anchor Network} \label{sec:supernet}
We introduce an anchor net $\mathrm{F}$ as an alternative approach to model weight generation in dynamic networks for MTL~\cite{navon2021learning,lin2020controllable}. Previous methods adopt chunking~\cite{ha2016hypernetworks} to mitigate the large computation and memory required for generating entire network weights at the expense of limiting the hypernet capacity. The anchor net, consisting of $N$-stream backbones trained for $N$ individual tasks (Figure~\ref{fig:framework}), overcomes this bottleneck by providing the weights in the tree structures predicted by the edge hypernet. Our choice of the anchor net is motivated by the need for an initialization that reflects inter-task relations and is based on observations from \cite{vandenhende2019branched}, where branching in tree-structured MTL networks is shown to be contingent on how similar task features are at any layer. It can also be interpreted as a supernet used in one-shot NAS approaches~\cite{bender2018understanding}, which is capable of emulating any architecture in the search space. Subsequently, the base weights of the anchor net are further modulated via the weight hypernet to address the cross-task connections unseen in the anchor net (Section~\ref{sec:weight_hypernet}).

\begin{figure}[t]
    \centering
    \includegraphics[width=0.35\textwidth]{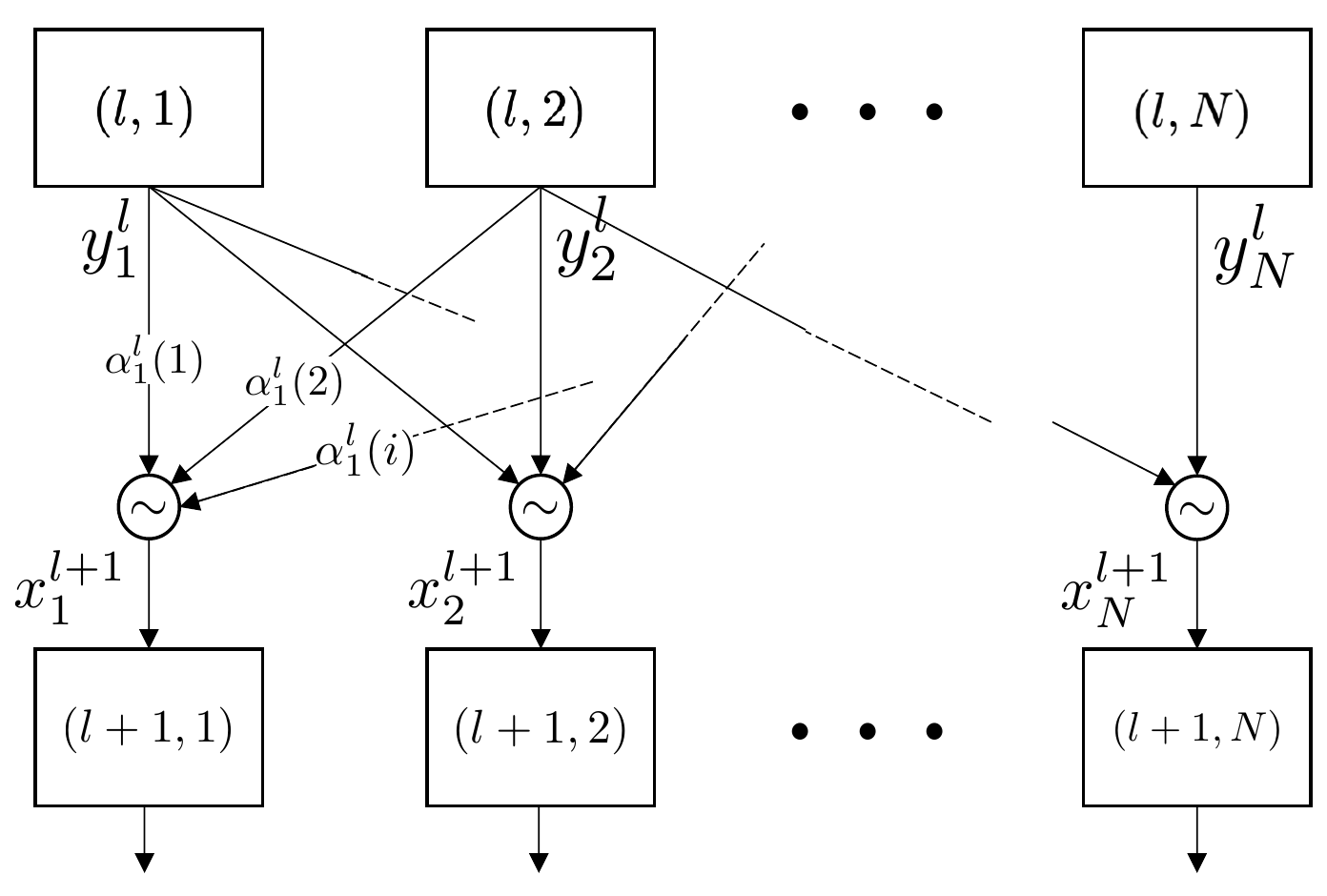}
    \vspace{-0.2cm}
    \caption{\textbf{Branching block.} Illustration of the parent sampling operation in Section \ref{sec:search_space}. Nodes in layer $l$ are sampled in accordance to a categorical distribution defined by $\alpha^l_j$ $\left(\sum_i \alpha^l_j(i)=1\right)$ for each node $(l+1,j)$ in layer $l+1$.}
    \label{fig:block}
    \vspace{-0.5cm}
\end{figure}

\subsection{Architecture Search Space} \label{sec:search_space}
We utilize a tree-structured network topology which has been shown to be highly effective for multi-task learning in \cite{guo2020learning}. It shares common low-level features over tasks while extracting task-specific ones in the higher layers, enabling control of the trade-off between tasks by changing branching locations conditioned on the desired preference $\left(\bm{r},c\right)$. The search space is represented as a \textit{directed acyclic graph} (DAG), where vertices in the graph represent different operations and edges denote the data flow through the network. Figure \ref{fig:block} shows a block of such a graph, containing $N$ \textit{parent} and \textit{child} nodes. In this work, we realize a tree-structure by stacking such blocks sequentially and allowing a child node to sample a path from the candidate paths between itself and all its parent nodes.
Concretely, we formulate the stochastic branching operation at layer $l$ as 
\begin{equation}
    x^{l+1}_j = d_j\cdot Y^l, \quad d_j\sim p_{\alpha_j^l},
\end{equation}
where $x^{l+1}_j$ denotes the input to the $j$-th node in layer $l+1$, $d_j$ is a one-hot vector indicating the parent node sampled from the \emph{categorical} distribution parameterized by $\alpha_j^l$ and, $Y^l = [ y^l_1, \dots, y^l_N ]$ concatenates outputs from all parent nodes at layer $l$.
Note that selecting a parent from every node determines a unique tree structure. This suggests learning $\bm{\alpha} = \{\alpha_j^l\}_{0 \leq j \leq N, 0 \leq l < L}$, 
conditioned on a preference $(\bm{r},c)$, in a manner which satisfies the desired task trade-offs. Here, $L$ denotes the total number of layers.





\subsection{Preference Conditioned Hypernetworks} \label{sec:hypernet}


%
We use two hypernets~\cite{ha2016hypernetworks} to construct our controller for architectural changes. The edge hypernet $h$, parameterized by $\phi$, predicts the branching parameters $\hat{\bm{\alpha}}=h(\bm{r},c; \phi)$ within the anchor net. Subsequently, the weight hypernet $\bar{h}$, parameterized by $\bar{\phi}$, predicts the normalization parameters $\{\hat{\bm{\beta}},\hat{\bm{\gamma}}\}=\bar{h}(\bm{r},c; \bar{\phi})$ to adapt the predicted network.

Optimizing the task loss $\mathcal{L}_{\textrm{task}}$ only takes into account the individual task performances without considering computational cost. 
Consequently, we introduce a branching regularizer $\Omega(\bm{r},c, \hat{\bm{\alpha}})$ to encourage node sharing (or branching) based on the preference. This regularizer contains two terms, the \emph{active} loss, which encourages limited sharing of features among the high preference tasks, and the \emph{inactive} loss, which aims to reduce resource utilization for the less important ones.
In particular, the active loss is additionally weighted by the cost preference $c$ to enable the control of total computational cost.
Formally, our objective is formulated as to find the controller ($\phi$ and $\bar{\phi}$) that minimizes the expectation of the branching regularized task loss over the distribution of user preferences $P_{(\bm{r},c)}$:
\begin{align}
    \min_{\phi, \bar{\phi}}
    \mathbb{E}_{(\bm{r}, c)\sim P_{(\bm{r},c)}}
    \left[\mathcal{L}_{\textrm{task}}(\bm{r}, \hat{\bm{\alpha}}, \hat{\bm{\beta}}, \hat{\bm{\gamma}}) 
    + \Omega(\bm{r}, c, \hat{\bm{\alpha}})\right],
\end{align}

We disentangle the training of the hypernetworks for stability -- the edge hypernet is trained first, followed by the weight hypernet.
At test time, when a preference $(\bm{r},c)$ is presented to the controller, the maximum likelihood architecture corresponding to the supplied preference is first sampled from the branching distribution parameterized by the predictions of $h$. The weights of this tree-structure are then inherited from the anchor net, supplemented via adapted normalization parameters predicted by $\bar{h}$.

\subsubsection{Regularizing the Edge Hypernet} \label{sec:edge_hypernet}
\begin{figure}[t]
    \centering
    \includegraphics[width=0.4\textwidth]{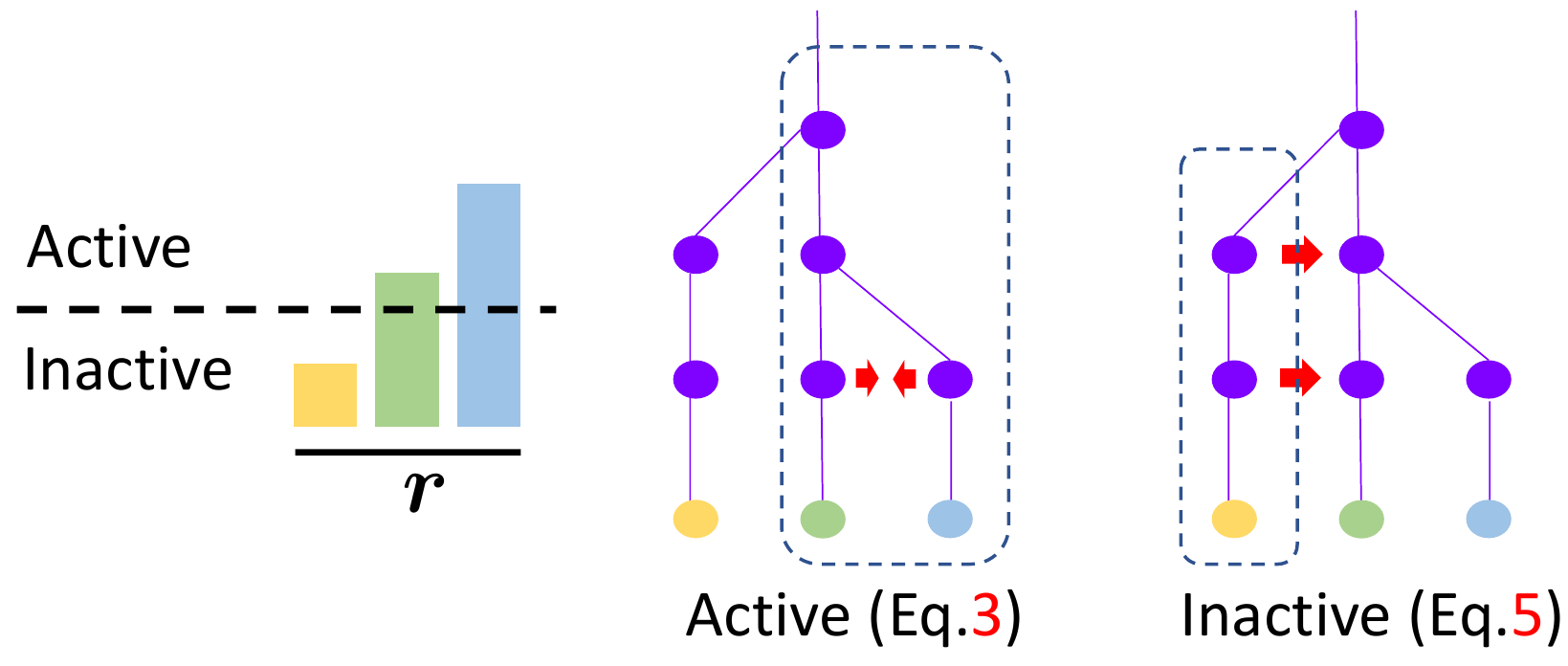}
    \vspace{-0.2cm}
    \caption{\textbf{Branching loss.} Illustration of the branching regularization, consisting of active and inactive losses. The active loss encourages limited sharing between high importance tasks, while the inactive loss tries to limit branching for less preferred tasks as much as possible.}
    \label{fig:branching_loss}
    \vspace{-0.5cm}
\end{figure}

We illustrate the idea of branching regularization in Figure~\ref{fig:branching_loss}: tasks with higher preferences should have a greater influence on the branching structure while tasks with smaller preferences may be de-emphasized by encouraging them to follow existing branching choices.
Specifically, we define two losses, active and inactive losses, based on the task division into two groups, active tasks $\mathcal{A}=\{\mathcal{T}_i \ | \ r_i \geq \tau, \ \forall i \in [N]\}$, and inactive tasks $\mathcal{I}=\{\mathcal{T}_i \ | \ r_i < \tau, \ \forall i \in [N]\}$ with some threshold $\tau$. 
Although individual tasks are already weighted by $\bm{r}$ in task loss $\mathcal{L}_{\textrm{task}}$, this explicit emphasizing of certain tasks over others was found to be crucial to induce better controllability, as shown in Section~\ref{sec:ablation}.\\
\noindent \textbf{Active loss.} The active loss $\mathcal{L}_{\textrm{active}}$ encourages nodes in the anchor net, 
corresponding to the active tasks, to be shared in order to avoid the whole network being split up by tasks with little knowledge shared among them.
Specifically, we encourage any pair of nodes that are likely to be sampled in the final architecture ($P$) and are from two similar tasks ($A$) to take the same parent node.
Formally, we define $\mathcal{L}_{\textrm{active}}$ as,
\begin{equation} \label{eq:active}
    \mathcal{L}_{\textrm{active}} = \sum_{l=1}^L \sum_{\substack{i,j \in \mathcal{A} \\ i\neq j}}
    \frac{L-l}{L} \cdot A(i,j)
    \cdot P(l,i,j) \cdot
    \|\nu^{l}_{i} - \nu^{l}_{j}\|^2,
\end{equation}
where $P(l,i,j) = P_{\textrm{use}}(l,i) \cdot P_{\textrm{use}}(l,j)$. $P_{use}(l,i)=1 - \prod_{k}\{ 1 - P_{use}(l+1,k) \cdot \nu^{l}_k(i) \}$ denotes the probability that the nodes $i$ in layer $l$ are used in the sampled tree structure. $A(i,j)$ captures the \emph{task affinity} between tasks $\mathcal{T}_i$ and $\mathcal{T}_j$, where we adopt \emph{Representational Similarity Analysis} (RSA)~\cite{dwivedi2019representation} to compute the affinity. The factor $\frac{L-l}{L}$ encourages more sharing of nodes which contain low-level features.
The full derivations of $P_{\textrm{use}}$ and $A$ are detailed in the supplementary document.

We use the Gumbel-Softmax reparameterization trick~\cite{jang2016categorical} to obtain the samples $\nu^l_i$ from the predicted logits $\hat{\bm{\alpha}}$,
\begin{equation} \label{eq:gumbel}
    \nu^{l}_{i}(k) = \frac{\exp \left((\log \alpha^l_i(k) + G^l_i(k))/\zeta\right)}{\sum_{m=1}^N\exp \left((\log \alpha^l_i(m) + G^l_i(m))/\zeta\right)}.
\end{equation}
Here, $G^l_i=-\log (-\log U^l_i)$ is a standard Gumbel distribution with $U^l_i$  sampled i.i.d. from the uniform distribution $\text{Unif}(0, 1)$, and $\zeta$ denotes the temperature of the softmax.

\noindent \textbf{Inactive loss.} 
The inactive tasks should have minimal effect in terms of branching. Inactive loss, $\mathcal{L}_{\textrm{inactive}}$, encourages these tasks to mimic the most closely related branching pattern, 
\begin{equation} \label{eq:inactive}
    \mathcal{L}_{\textrm{inactive}} = \sum_{l=1}^{L} \sum_{j \in \mathcal{I}}
     \min_{i \in \mathcal{A}} \|\nu^{l}_{i} - \nu^{l}_{j}\|^2.
\end{equation}
This ensures that the network branching is controlled by the active tasks, with the inactive tasks sharing nodes with the active tasks.

Thus, the branching regularizer is defined as follows,
\begin{equation}
    \Omega(\bm{r}, c, \hat{\bm{\alpha}}) = c\cdot\lambda_{\mathcal{A}}\mathcal{L}_{\textrm{active}} + \lambda_{\mathcal{I}}\mathcal{L}_{\textrm{inactive}},
\end{equation}
where $\lambda_{\mathcal{A}}$, $\lambda_{\mathcal{I}}$ are hyperparameters to determine the weighting of the losses. Typically, we set $\lambda_{\mathcal{A}}=1$ and $\lambda_{\mathcal{I}}=0.1$.
Here, the active loss is additionally weighted by the resource preference $c$, so that larger $c$ encourages more feature sharing to reduce total computational cost.

\subsubsection{Cross-task Adaptation} \label{sec:weight_hypernet}
The architecture sampled by the edge hypernet $h$ contains edges that have not been observed during the anchor net training. These are denoted as \emph{cross-task} edges since they connect nodes that belong to different streams in $\mathrm{F}$. 
Consequently, the performance of the sampled network is sub-optimal. To rectify this issue, we propose to modulate the weights of the anchor net to adaptively update the unseen edges using an additional weight hypernet $\bar{h}$. Inspired from the prior works~\cite{wang2020tent,mahabadi2021parameter} that estimate normalization statistics and optimize channel-wise affine transformations, we modulate only the normalization parameters using a hypernetwork. Concretely, we modulate the original batch normalization operation at layer $l$,
$\text{BN}_i^l(\bm{x}^l_i) = \gamma^l_i\frac{x^l_i-\mu^l_{i}}{\sigma_i^l} + \beta^l_i$, to 
$\text{BN}_i^l(\bm{x}^l_i) = (\gamma^l_i+\Delta\bm{\gamma^l_i})\frac{x^l_i-\mu^l_{i}}{\sigma_i^l} + (\beta^l_i+\Delta\bm{\beta^l_i})$
by predicting the perturbations to the parameters: $\{\Delta \beta^l_i,\Delta \gamma^l_i\}_{0 \leq i \leq N, 0 \leq l < L} = \bar{h}(r,c;\theta)$,
where $\gamma^l_i$ and $\beta^l_i$ are the original affine parameters, and $\mu^l_i$ and $\sigma^l_i$ denote the batch statistics of the node input $x^l_i$.
This modulation primarily affects the preferences with two or more dominant tasks, where cross-task connections occur.



\section{Experiments}
In this section, we demonstrate the ability of our framework to dynamically search for efficient architectures for multi-task learning. We show that our framework achieves flexibility between two extremes of the accuracy-efficiency trade-off, allowing a better control within a single model. Extensive experiments indicate that the predicted network structures match well with the input preferences, in terms of both resource usage and task performance.
\subsection{Evaluation Criteria}
\noindent \textbf{Uniformity.}
To measure controllability with respect to task preferences, we utilize uniformity~\cite{mahapatra2020multi} which quantifies how well the vector of task losses $\bm{L} = [\mathcal{L}_1,\dots,\mathcal{L}_N]$ is aligned with the given preference. Specifically, for the loss vector $\bm{L}$ corresponding to the architecture for task preference $\bm{r}$, uniformity is defined as $\mu_{\bm{r}} = 1 - \infdiv{\hat{\bm{L}}}{\bm{1}/N}$, where $\hat{\bm{L}}(j) = \frac{r_j\mathcal{L}_j}{\sum_i r_i\mathcal{L}_i}$. This arises from the fact that, ideally, $r_j \propto 1/\mathcal{L}_j$, which in turn implies $r_1\mathcal{L}_1=r_2\mathcal{L}_2\dots=r_N\mathcal{L}_N$. 

\noindent \textbf{Hypervolume.}
Using the trained controller, we are able to approximate the trade-off curve among the different tasks in the loss space. To evaluate the quality of this curve we utilize hypervolume (HV)~\cite{zitzler1999multiobjective} -- a popular metric in the multi-objective optimization literature to compare different sets of solutions approximating the Pareto front \cite{fleischer2003measure}. It measures the volume in the loss space of points dominated by a solution in the evaluated set. Since this volume is unbounded, hypervolume measures the volume in a rectangle defined by the solutions and a selected \emph{reference point}. More details can be found in the appendix.

\noindent \textbf{Computational Resource.}
We measure the computational cost using the memory of the activated nodes in the anchor net and the GFLOPs, which approximates the time spent in the forward pass.
We also report the computational cost of the hypernets to take into account their overheads. 
We discuss more on the model size in the appendix.
\subsection{Datasets}
We evaluate the performance of our approach using three multi-task datasets, namely \textbf{PASCAL-Context}~\cite{mottaghi2014the} and \textbf{NYU-v2}~\cite{silberman2012indoor}, and \textbf{CIFAR-100}~\cite{krizhevsky2009learning}. The PASCAL-Context dataset is used for joint semantic segmentation, human parts segmentation and saliency estimation, as well as these three tasks together with surface normal estimation, and edge detection as in \cite{bruggemann2020automated}. The NYU-v2 dataset comprises images of indoor scenes, fully labeled for semantic segmentation, depth estimation and surface normal estimation. For CIFAR-100, we split the dataset into 20 five-way classification tasks~\cite{rosenbaum2017routing}.
\subsection{Baselines}
We compare our framework with both \emph{static} and \emph{dynamic} networks. Static networks include \textbf{Single-task} networks, where we train each task separately using a task-specific backbone, and \textbf{Multi-task} networks, in which all tasks share the backbone but have separate task-specific heads at the end. These multi-task networks are trained separately for different preferences and thus, training time scales linearly with the number of preferences. We use this to contrast the training time of our framework. The single-task networks demonstrate the anchor net performance. We also compare our architectures with two multi-task NAS methods, \textbf{LTB}~\cite{li2020learning} and \textbf{BMTAS}~\cite{bruggemann2020automated}, which use the same tree-structured search space to perform NAS, but are static. The dynamic networks include Pareto Hypernetworks (\textbf{PHN}) \cite{navon2021learning}, which  predicts only the weights of a shared backbone network conditioned on a task preference vector using hypernetworks, and \textbf{PHN-BN}, a variation of PHN which predicts only the normalization parameters similar to our weight hypernet. Implementation details are presented in the appendix.
%
%
%

\begin{figure}[t]
    \centering
    \includegraphics[width=0.4\textwidth]{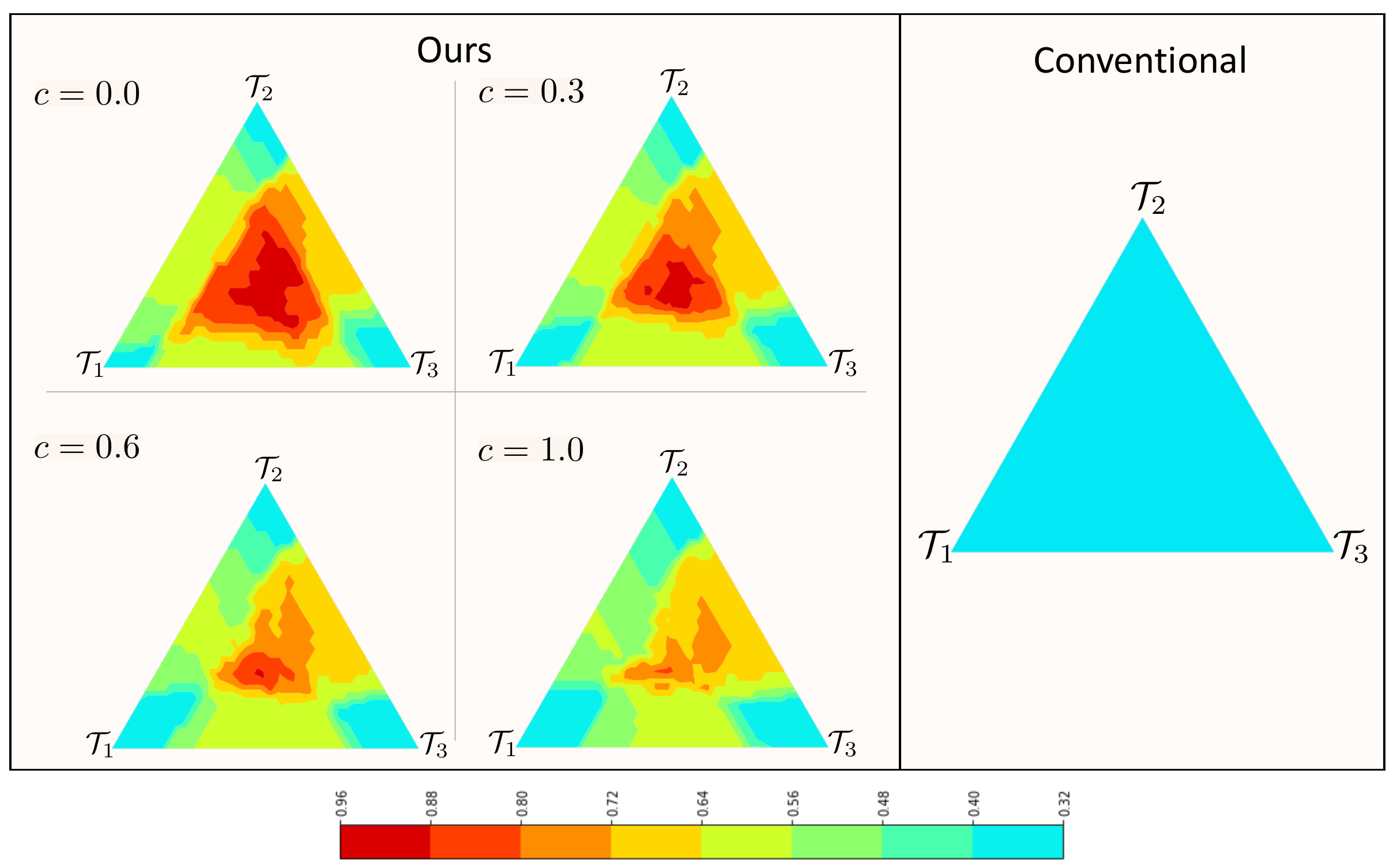}
    \vspace{-0.25cm}
    \caption{\textbf{Resource usage on NYU-v2.} We visualize resource usage by plotting the proportion of parameters active in the anchor net versus the task preference. The three vertices represent the task preferences with non-zero importance to only one task, while areas in the middle correspond to more dense preferences. As $c$ increases, the predicted networks grow progressively smaller in the dense regions. On the other hand, conventional dynamic networks for MTL always have a constant resource usage ($\mathcal{T}_1$:semantic seg., $\mathcal{T}_2$:surface normal, $\mathcal{T}_3$:depth).}
    \label{fig:resource}
    \vspace{-0.5cm}
\end{figure}

\subsection{Comparison with Baselines}

\noindent \textbf{Controllable resource usage.}
We visualize the variation in computational cost with respect to different task and resource usage preferences in Figure~\ref{fig:resource}.
We adopt the ratio of the size of the predicted architecture to the size of the total anchor net as the criterion for evaluating computational cost. 
Compared to conventional dynamic networks that only adjust weights with a fixed computational cost (right), our framework (left) enables control over the total cost via a cost preference $c$. 
Resource usage peaks at the center of the contour, when more tasks are active, and falls down gradually as we move towards the corners, where task preferences are heavily skewed. Furthermore, the average resource usage decreases as $c$ is increased, indicating the ability of the controller to incorporate resource constraints.

\noindent \textbf{Multi-task performance.}
We demonstrate the overall multi-task performance in Tables~\ref{tab:pascal5_perf}-\ref{tab:cifar_perf} on four different 
settings (PASCAL-Context 5-task, PASCAL-Context 3-task, NYU-v2 3-task, CIFAR-100 20-task). In all cases, we report hypervolume (reference point mentioned below heading) and uniformity averaged across $20$ task preference vectors $\bm{r}$, sampled uniformly from $\mathcal{S}_N$. Inference network cost is calculated similarly
over $1000$ preference vectors. These are shown for two choices of $c\in \{0,1\}$ to highlight the two extreme cases of resource usage. 

Our framework achieves higher values in both hypervolume and uniformity compared to the existing dynamic models (PHN and PHN-BN) in all four settings.
While the high hypervolume reinforces the efficacy of tree-structured models in solving multi-task problems, the uniformity values consolidate architectural change as an effective approach towards modeling task trade-offs.
This is accompanied by increased average computational cost, indicated by inference parameter count.
As discussed above, this is due to the flexible architecture over preferences, where actual cost will differ for each preference, \eg, reaching the cost of PHN-BN for extremely skewed preferences (Figure~\ref{fig:resource}). 
Compared to Single-Task, the proposed controller is able to find effective architectures (as indicated by the hypervolume) which perform nearly at par 
with a smaller memory footprint (as indicated by the average inference network parameter count). Notably, in the NYU-v2 3-task and CIFAR-100 settings,
the ability to find effective architectures enables the model to
outperform single-task networks, demonstrating the benefit of sharing features among related tasks via architectural change.
In addition, our framework enjoys flexibility between two extreme cases, \ie Single-Task (highest accuracy with lowest inference efficiency) and dynamic models with shared backbone (lowest accuracy with highest inference efficiency), spanning a range of trade-offs for different $c$ values. The range of HV is larger when task-specific features are useful, compared to when the compact architecture already achieves higher HV than the Single-Task (Tables~\ref{tab:nyu_perf},\ref{tab:cifar_perf}).
``Control Params.'' is the cost of the hypernets. Note that this overhead will materialize only when the preference changes and does not have any effect on the task inference time.

\begin{table}[t] 

\centering
\resizebox{\columnwidth}{!}{
\begin{tabular}{@{}lLcLLL@{}}
\toprule
\multicolumn{1}{l}{\textbf{Method}} & \textbf{HV.}$\uparrow$ $[3,3,\dots,3]$    & \textbf{Unif.}$\uparrow$ & \textbf{Inference Params.}$\downarrow$ & \textbf{GFLOPs}$\downarrow$ & \textbf{Control Params.} \\  \midrule
Single-Task                & 81.56  & -     &  9.84M    & 16.17  & -                                                                    \\ \midrule
PHN                        & 42.61 & 0.72   & 2.15M & 6.28 & 21.50M                                                                        \\
PHN-BN                     & 72.27 & 0.69   & 2.15M & 6.28 & 3.63M                                                                     \\ \midrule
Ours w/o adaptation, c=0.0                & 47.73 & 0.84  & 3.34M & 7.21 & 0.06M                                                                       \\
Ours w/o adaptation, c=1.0                & 30.91 & \textbf{0.86}  & 2.75M & 6.81 & 0.06M                                                                        \\
Ours, c=0.0  & \textbf{75.52} & 0.76  & 3.34M & 7.21 & 15.32M                                                                   \\
Ours, c=1.0  & 73.20 & 0.79  & 2.75M & 6.81  & 15.32M                                                                     \\ \bottomrule
\end{tabular}}
\vspace{-0.25cm}
\caption{\textbf{Evaluation on PASCAL-Context (5 tasks).}}
\label{tab:pascal5_perf}
\vspace{-0.3cm}
\end{table}
\begin{table}[t] 
\centering
\resizebox{\columnwidth}{!}{
\begin{tabular}{@{}lLcLLL@{}}
\toprule
\multicolumn{1}{l}{\textbf{Method}} & \textbf{HV.}$\uparrow$ $[3,3,3]$    & \textbf{Unif.}$\uparrow$ & \textbf{Inference Params.}$\downarrow$ & \textbf{GFLOPs}$\downarrow$ & \textbf{Control Params.} \\ \midrule
Single-Task                & 4.31 & -     & 5.91M &9.75 & -                                                                        \\ \midrule
PHN                        & 1.97 & 0.74  & 2.06M &4.81 & 21.10M                                                                        \\
PHN-BN                     & 3.92 & 0.79  & 2.06M &4.81 & 3.32M                                                                        \\ \midrule
Ours w/o adaptation, c=0.0                & 3.56 & \textbf{0.92}  & 3.15M  &5.52 & 0.03M                                                                    \\
Ours w/o adaptation, c=1.0                & 3.35 & 0.91  & 2.86M &5.07 & 0.03M                                                                       \\ 
Ours, c=0.0  & \textbf{4.26} & 0.82  & 3.15M &5.52 & 9.25M                                                                       \\
Ours, c=1.0  & 4.25 & 0.82  & 2.86M &5.07 & 9.25M                                                                      \\ \bottomrule
\end{tabular}}
\vspace{-0.25cm}
\caption{\textbf{Evaluation on PASCAL-Context (3 tasks).}}
\label{tab:pascal3_perf}
\vspace{-0.3cm}
\end{table}
\begin{table}[t!]
\centering
\resizebox{\columnwidth}{!}{
\begin{tabular}{@{}lLcLLL@{}}
\toprule
\multicolumn{1}{l}{\textbf{Method}} & \textbf{HV.}$\uparrow$ $[4,4,4]$   & \textbf{Unif.}$\uparrow$ & \textbf{Inference Params.}$\downarrow$ & \textbf{GFLOPs}$\downarrow$ & \textbf{Control Params.} \\ \midrule
Single-Task                & 12.83 & -     & 64.47M &58.78 & -                                                                        \\ \midrule
PHN                        & 2.36  & 0.75  & 21.59M  &21.02 &21.04M                                                                     \\
PHN-BN                     & 11.72 & 0.73  & 21.59M   &21.02 &2.23M                                                                    \\ \midrule
Ours w/o adaptation, c=0.0                & 12.42 & 0.82  & 41.06M &29.04 &0.03M                                                                        \\
Ours w/o adaptation, c=1.0                & 9.53  & \textbf{0.84}  & 34.68M &25.98 &0.03M                                                                        \\
Ours, c=0.0  & \textbf{13.43} & 0.76  & 41.06M &29.04 &5.72M                                                                      \\
Ours, c=1.0  & 13.08 & 0.78  & 34.68M &25.98 &5.72M                                                                       \\ \bottomrule
\end{tabular}}
\vspace{-0.25cm}
\caption{\textbf{Evaluation on NYU-v2}.}
\label{tab:nyu_perf}
\vspace{-0.3cm}
\end{table}
\begin{table}[t]
\centering
\resizebox{\columnwidth}{!}{
\begin{tabular}{@{}lLcLLL@{}}
\toprule
\multicolumn{1}{l}{\textbf{Method}} & \textbf{HV.}$\uparrow$ $[1,1,\dots,1]$    & \textbf{Unif.}$\uparrow$ & \textbf{Inference Params.}$\downarrow$ & \textbf{GFLOPs}$\downarrow$  & \textbf{Control Params.} \\ \midrule
Single-Task                & 0.009 & -     & 36.18M &348.79 &-                                                                        \\ \midrule
PHN                        & 0.002  & 0.54  & 16.35M  &73.13 &11.03M                                                                     \\
PHN-BN                     & 0.007 & 0.49  & 16.35M   &73.13 &0.31M                                                                    \\ \midrule
Ours w/o adaptation, c=0.0                & 0.003 & \textbf{0.58}  & 31.86M &174.36 &0.34M                                                                        \\
Ours w/o adaptation, c=1.0                & 0.001  & 0.53  & 31.37M &129.23 &0.34M                                                                        \\
Ours, c=0.0  & \textbf{0.010} & 0.54  & 31.86M &174.36 &3.10M                                                                      \\
Ours, c=1.0  & 0.009 & 0.49  & 31.37M &129.23 &3.10M                                                                       \\ \bottomrule
\end{tabular}}
\vspace{-0.25cm}
\caption{\textbf{Evaluation on CIFAR-100}.}
\label{tab:cifar_perf}
\vspace{-0.6cm}
\end{table}

\noindent \textbf{Effect of cross-task adaptation}
In Tables~\ref{tab:pascal5_perf}-\ref{tab:cifar_perf}, 
``Ours w/o adaptation'' denotes the model without weight hypernet. As indicated by larger HVs, cross-task adaptation improves the performance without affecting the inference time.
A trend that persists across all the settings is the slight drop in uniformity that accompanies the adapted models in comparison to the unadapted ones. This is due to the propensity of the weight hypernet to improve task performance as much as possible while keeping the preferences intact. This leads to improved performance even in the low-priority tasks at the expense of lower uniformity. Note that our primary factor of controllability is through architectural changes which remains unaffected by the weight hypernet.

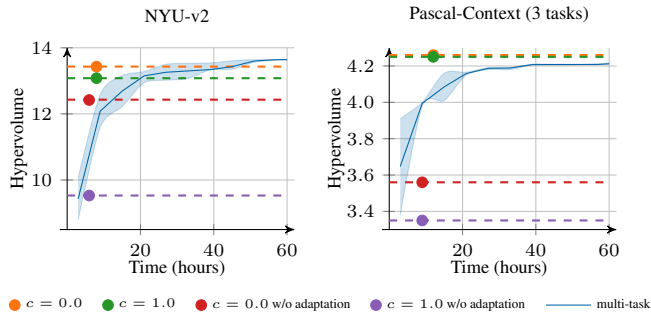
\begin{figure}\centering
\begin{tikzpicture}[
    >=stealth',
    tight background,
]
\begin{axis}[
    width=4.7cm,
    height=4cm,
    grid=major,
    major grid style={line width=.2pt,draw=gray!50},
    xmin=-2,
    xmax=62,
    ymin=8.5,
    ymax=14.0,
    axis lines=middle,
    axis line style={->},
    x label style={at={(axis description cs:0.5,-0.12)},anchor=north},
    y label style={at={(axis description cs:-0.1,.5)},rotate=90,anchor=south},
    y tick label style={
    /pgf/number format/.cd,
    fixed,
    fixed zerofill,
    precision=0,
    /tikz/.cd
    },
    xlabel={Time (hours)},
    ylabel={Hypervolume},
    legend style={font=\tiny,text=black,draw=none,/tikz/every even column/.append style={column sep=0.2cm}},
    legend columns=-1,
    legend to name=named,
    title={NYU-v2},
]
\addplot[only marks,mark=*,mark size=2pt,plotsorange] coordinates {
    (8.0, 13.43)
};
\addlegendentry{$c=0.0$};
\addplot[only marks,mark=*,mark size=2pt,plotsgreen] coordinates {
    (8.0, 13.08)
};
\addlegendentry{$c=1.0$};
\addplot[only marks,mark=*,mark size=2pt,plotsred] coordinates {
    (6.0, 12.42)
};
\addlegendentry{$c=0.0$ w/o adaptation};
\addplot[only marks,mark=*,mark size=2pt,plotspurple] coordinates {
    (6.0, 9.53)
};
\addlegendentry{$c=1.0$ w/o adaptation};

\addplot[color=plotsorange,dashed,thick,forget plot] coordinates {
    (0.0, 13.43)
    (60.0, 13.43)
};
\addplot[color=plotsgreen,dashed,thick,forget plot] coordinates {
    (0.0, 13.08)
    (60.0, 13.08)
};
\addplot[color=plotsred,dashed,thick,forget plot] coordinates {
    (0.0, 12.43)
    (60.0, 12.43)
};
\addplot[color=plotspurple,dashed,thick, forget plot] coordinates {
    (0.0, 9.53)
    (60.0, 9.53)
};

\addplot [no marks,plotsblue,smooth,name path=A,opacity=0.2,forget plot] plot coordinates {
   (3, 10.06)
   (9, 12.58)
   (15, 13.16)
   (21, 13.28)
   (27, 13.50)
   (33, 13.53)
   (39, 13.54)
   (45, 13.61)
   (51, 13.64)
   (57, 13.64)
   (60, 13.64)
};

\addplot [no marks,plotsblue,smooth,name path=B,opacity=0.2,forget plot] plot coordinates {
   (3, 8.79)
   (9, 11.58)
   (15, 12.22)
   (21, 12.94)
   (27, 13.02)
   (33, 13.07)
   (39, 13.33)
   (45, 13.35)
   (51, 13.56)
   (57, 13.63)
   (60, 13.64)
};

\addplot [no marks,plotsblue,name path=C] 
plot coordinates {
   (3, 9.43)
   (9, 12.08)
   (15, 12.69)
   (21, 13.15)
   (27, 13.26)
   (33, 13.30)
   (39, 13.34)
   (45, 13.43)
   (51, 13.60)
   (57, 13.64)
   (60, 13.64)
};

\addlegendentry{multi-task};
\addplot[plotsblue,forget plot,fill opacity=0.2] fill between[of=A and B];

\end{axis}
\end{tikzpicture}
\hfill
\begin{tikzpicture}[
    >=stealth',
    tight background,
]
\begin{axis}[
    width=4.7cm,
    height=4cm,
    grid=major,
    major grid style={line width=.2pt,draw=gray!50},
    xmin=-2,
    xmax=62,
    ymin=3.3,
    ymax=4.3,
    axis lines=middle,
    axis line style={->},
    x label style={at={(axis description cs:0.5,-0.12)},anchor=north},
    y label style={at={(axis description cs:-0.12,.5)},rotate=90,anchor=south},
    y tick label style={
    /pgf/number format/.cd,
    fixed,
    fixed zerofill,
    precision=1,
    /tikz/.cd
    },
    xlabel={Time (hours)},
    ylabel={Hypervolume},
    title={Pascal-Context (3 tasks)},
]
\addplot[only marks,mark=*,mark size=2pt,plotsorange] coordinates {
    (12.0, 4.26)
};
\addplot[only marks,mark=*,mark size=2pt,plotsgreen] coordinates {
    (12.0, 4.25)
};
\addplot[only marks,mark=*,mark size=2pt,plotsred] coordinates {
    (9.0, 3.56)
};
\addplot[only marks,mark=*,mark size=2pt,plotspurple] coordinates {
    (9.0, 3.35)
};

\addplot[color=plotsorange,dashed,thick,forget plot] coordinates {
    (0.0, 4.26)
    (60.0, 4.26)
};
\addplot[color=plotsgreen,dashed,thick,forget plot] coordinates {
    (0.0, 4.25)
    (60.0, 4.25)
};
\addplot[color=plotsred,dashed,thick,forget plot] coordinates {
    (0.0, 3.56)
    (60.0, 3.56)
};
\addplot[color=plotspurple,dashed,thick, forget plot] coordinates {
    (0.0, 3.35)
    (60.0, 3.35)
};

\addplot [no marks,plotsblue,smooth,name path=A,opacity=0.2,forget plot] plot coordinates {
   (3, 3.91)
   (9, 4.00)
   (15, 4.16)
   (21, 4.166)
   (27, 4.19)
   (33, 4.19)
   (39, 4.21)
   (45, 4.21)
   (51, 4.21)
   (57, 4.211)
   (60, 4.218)
};
\addplot [no marks,plotsblue,smooth,name path=B,opacity=0.2,forget plot] plot coordinates {
   (3, 3.38)
   (9, 3.98)
   (15, 4.01)
   (21, 4.15)
   (27, 4.18)
   (33, 4.18)
   (39, 4.205)
   (45, 4.205)
   (51, 4.206)
   (57, 4.206)
   (60, 4.21)
};

\addplot [no marks,plotsblue,name path=C] 
plot coordinates {
   (3, 3.647)
   (9, 3.994)
   (15, 4.085)
   (21, 4.158)
   (27, 4.187)
   (33, 4.189)
   (39, 4.2074)
   (45, 4.2079)
   (51, 4.2079)
   (57, 4.2086)
   (60, 4.212)
};
\addplot[plotsblue,forget plot,fill opacity=0.2] fill between[of=A and B];
\end{axis}
\end{tikzpicture}
\ref{named}
\vspace{-0.7cm}
\caption{\textbf{Comparison with preference-specific multi-task networks.} For static multi-task models, each value is computed by evaluating a subset of preferences, with the shaded area marking the variance across selected subsets. Our framework achieves high hypervolume significantly faster with a single model.}
\vspace{-0.5cm}
\label{fig:multitask_hv}
\end{figure}

\noindent \textbf{Training efficiency.}
In contrast to dynamic networks, static multi-task networks require multiple models to be trained, corresponding to different task preferences, to approximate the trade-off curve. As a result, these methods have a clear trade-off between their performance and their training time. To analyze this trade-off, we plot hypervolume vs. training time for our framework when compared to training multiple static models in Figure \ref{fig:multitask_hv}. We trained 20 multi-task models with different preferences sampled uniformly, and at the inference time we selected subsets of various sizes and computed their hypervolume. The shaded area in Figure~\ref{fig:multitask_hv} reflects the variance over different selections of task preference subsets.
This empirically shows that our approach requires shorter training time to achieve similar hypervolume compared to the static multi-task networks.
\begin{figure}\centering%
\begin{tikzpicture}[>=stealth',mark size=1.5pt]
\begin{axis}[
    width=4.5cm,
    height=3cm,
    grid=major,
    major grid style={line width=.2pt,draw=gray!50},
    xmin=0,
    xmax=1.0,
    ymin=1.35,
    ymax=3.25,
    axis lines=middle,
    axis line style={->},
    x label style={at={(axis description cs:0.5,-0.225)},anchor=north},
    y label style={at={(axis description cs:-0.175,.5)},rotate=90,anchor=south},
    y tick label style={
        /pgf/number format/.cd,
        fixed,
        fixed zerofill,
        precision=1,
        /tikz/.cd
    },
    xlabel={Task preference},
    ylabel={Loss},
    legend style={at={(0.45, 0.8)}, anchor=west, draw=none},
    legend cell align={left},
    title={Semantic Segmentation},
    every axis title/.style={anchor=south, at={(0.5,1.0)}},
]
\addplot[color=plotsorange,solid,thick,mark=*, mark options={fill=white,solid}] coordinates {
    (0.1, 3.15)
    (0.3, 2.22)
    (0.5, 1.69)
    (0.7, 1.52)
    (0.9, 1.51)
};
\addplot[color=plotsblue,solid,thick,mark=*, mark options={fill=white,solid}] coordinates {
    (0.1, 2.94)
    (0.3, 1.65)
    (0.5, 1.57)
    (0.7, 1.52)
    (0.9, 1.51)
};
\addplot[color=plotsorange,dashed,thick,mark=*, mark options={fill=white,solid}] coordinates {
    (0.1, 1.86)
    (0.3, 1.50)
    (0.5, 1.59)
    (0.7, 1.62)
    (0.9, 1.62)
};
\addplot[color=plotsblue,dashed,thick,mark=*, mark options={fill=white,solid}] coordinates {
    (0.1, 1.72)
    (0.3, 1.54)
    (0.5, 1.60)
    (0.7, 1.63)
    (0.9, 1.60)
};
\end{axis}
\end{tikzpicture} 
\hfill
\begin{tikzpicture}[>=stealth',mark size=1.5pt]
\begin{axis}[
    width=4.5cm,
    height=3cm,
    grid=major,
    major grid style={line width=.2pt,draw=gray!50},
    xmin=0,
    xmax=1.0,
    ymin=1.9,
    ymax=4.5,
    axis lines=middle,
    axis line style={->},
    x label style={at={(axis description cs:0.5,-0.225)},anchor=north},
    y label style={at={(axis description cs:-0.175,.5)},rotate=90,anchor=south},
    y tick label style={
        /pgf/number format/.cd,
        fixed,
        fixed zerofill,
        precision=1,
        /tikz/.cd
    },
    xlabel={Task preference},
    ylabel={Loss},
    legend style={at={(0.45, 0.8)}, anchor=west, draw=none},
    legend cell align={left},
    title={Surface Normals},
    every axis title/.style={anchor=south, at={(0.5,1.0)}},
]
\addplot[color=plotsorange,solid,thick,mark=*, mark options={fill=white,solid}] coordinates {
    (0.1, 4.32)
    (0.3, 2.81)
    (0.5, 2.28)
    (0.7, 2.11)
    (0.9, 2.06)
};
\addplot[color=plotsblue,solid,thick,mark=*, mark options={fill=white,solid}] coordinates {
    (0.1, 4.27)
    (0.3, 2.31)
    (0.5, 2.18)
    (0.7, 2.12)
    (0.9, 2.06)
};
\addplot[color=plotsorange,dashed,thick,mark=*, mark options={fill=white,solid}] coordinates {
    (0.1, 3.11)
    (0.3, 2.48)
    (0.5, 2.09)
    (0.7, 2.07)
    (0.9, 2.05)
};
\addplot[color=plotsblue,dashed,thick,mark=*, mark options={fill=white,solid}] coordinates {
    (0.1, 3.33)
    (0.3, 2.32)
    (0.5, 2.06)
    (0.7, 2.05)
    (0.9, 2.04)
};
\end{axis}
\end{tikzpicture} 
\hfill
\begin{tikzpicture}[>=stealth',mark size=1.5pt]
\begin{axis}[
    width=4.5cm,
    height=3cm,
    grid=major,
    major grid style={line width=.2pt,draw=gray!50},
    xmin=0,
    xmax=1.0,
    ymin=1.15,
    ymax=2.9,
    axis lines=middle,
    axis line style={->},
    x label style={at={(axis description cs:0.5,-0.225)},anchor=north},
    y label style={at={(axis description cs:-0.175,.5)},rotate=90,anchor=south},
    y tick label style={
        /pgf/number format/.cd,
        fixed,
        fixed zerofill,
        precision=1,
        /tikz/.cd
    },
    xlabel={Task preference},
    ylabel={Loss},
    title={Depth},
    every axis title/.style={anchor=south, at={(0.5,1.0)}},
    legend style={
        anchor=north west,
        at={(1.25,1.0)},
        text=black,
        draw=none,
        /tikz/every even column/.append style={column sep=0.2cm},
    },
    legend cell align={left},
]
\addplot[color=plotsblue,solid,thick,mark=*, mark options={fill=white,solid}] coordinates {
    (0.1, 2.52)
    (0.3, 1.39)
    (0.5, 1.36)
    (0.7, 1.33)
    (0.9, 1.32)
};
\addlegendentry{$c=0.0$ w/o adaptation};
\addplot[color=plotsorange,solid,thick,mark=*, mark options={fill=white,solid}] coordinates {
    (0.1, 2.87)
    (0.3, 1.43)
    (0.5, 1.38)
    (0.7, 1.33)
    (0.9, 1.32)
};
\addlegendentry{$c=1.0$ w/o adaptation} ;
\addplot[color=plotsblue,dashed,thick,mark=*, mark options={fill=white,solid}] coordinates {
    (0.1, 1.75)
    (0.3, 1.31)
    (0.5, 1.30)
    (0.7, 1.29)
    (0.9, 1.28)
};
\addlegendentry{$c=0.0$};
\addplot[color=plotsorange,dashed,thick,mark=*, mark options={fill=white,solid}] coordinates {
    (0.1, 1.82)
    (0.3, 1.30)
    (0.5, 1.30)
    (0.7, 1.29)
    (0.9, 1.29)
};
\addlegendentry{$c=1.0$};
\end{axis}
\end{tikzpicture} 
\hfill
\vspace{-0.3cm}
\caption{\textbf{Marginal evaluation tasks on NYU-v2}}
\label{fig:marginal_nyu}
\vspace{-0.6cm}
\end{figure}
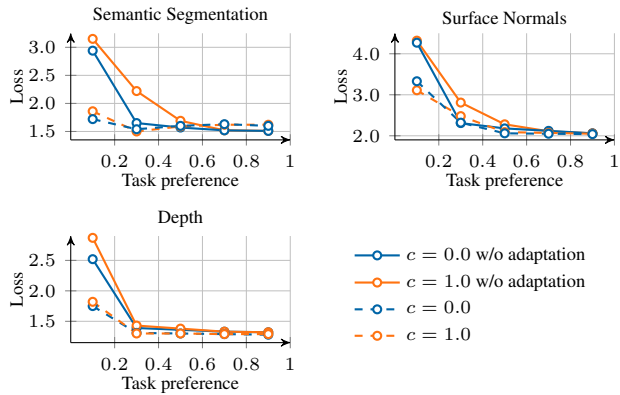
\begin{table}[]
\centering
\resizebox{\columnwidth}{!}{
\begin{tabular}{@{}lccccc@{}}
\toprule
\textbf{Method} & $\mathcal{T}_1$ $\uparrow$ & $\mathcal{T}_2$ $\uparrow$ & $\mathcal{T}_3$ $\uparrow$ & Avg $\Delta_{\mathbf{T}} (\%) \uparrow$ & \# \textbf{Params} (\%) $\downarrow$                  \\ \midrule
Single-Task                & 64.11   & 58.41  & 65.17  & -    & -    \\ \midrule
LTB                        & 61.84   & 59.41  & 64.18  & -1.12    & -35.0    \\
BMTAS                      & 62.79   & 58.41  & 64.74  & -0.93    & \textbf{-48.9}    \\ \midrule
Ours, c=0.0                & 63.60                    & 59.41                & 64.94                  & \textbf{+0.18}                     & -35.2                     \\ 
$\text{Ours}^\dagger$, c=0.0                & 62.34                 & 58.60                  & 65.17                  & -0.81                    & -35.2                     \\
Ours, c=1.0                & 63.12                 & 58.93                    & 64.93                   & -0.34                    & -40.8 \\

$\text{Ours}^\dagger$, c=1.0                & 61.91                 & 58.71                 & 65.01                     & -1.05                    & -40.8                    \\ \bottomrule
\end{tabular}}
\vspace{-0.15cm}
\caption{\textbf{Architecture evaluation on PASCAL-Context (3 tasks).} We report the mean intersection over union for $\mathcal{T}_1:\text{Semantic seg.}$, $\mathcal{T}_2:\text{Parts seg.}$, and $\mathcal{T}_3:\text{Saliency}$. Presence of $\dagger$ indicates that we train the networks initialized from ImageNet weights, while its absence indicates training from anchor net weights.} \label{tab:nyu_arch}
\vspace{-0.5cm}
\end{table}
\subsection{Analysis}

\noindent \textbf{Architecture evaluation.}
We study the effectiveness of the architectures predicted by the edge hypernet by comparing them with those predicted by LTB~\cite{guo2020learning} and BMTAS~\cite{bruggemann2020automated}. We choose the architecture predicted for a uniform task preference and, similar to LTB, we retrain it for a fair comparison. We evaluate the performance in terms of relative drop in performance across tasks and number of parameters with respect to the single task baseline. Despite not being directly trained for NAS, our framework is able to output architectures which perform at par with LTB (Table~\ref{tab:nyu_arch}). More results on the Pascal-Context dataset and visualization of dynamic architectures can be found in the appendix.

\noindent \textbf{Task controllability.}
In Figure \ref{fig:marginal_nyu} we visualize the task controllability for our framework by plotting the test loss at different values of task preference for that specific task, marginalized over preference values of the other tasks. As expected, increasing the preference for a task gradually leads to a decrease in the loss value. Furthermore, increasing $c$ leads to higher loss values due to smaller predicted architectures. The effect of the weight hypernet is also evident, as shown by the lower loss values obtained on using it on top of the edge hypernet (w/o adaptation).



\subsection{Ablation Study} \label{sec:ablation}
\noindent \textbf{Impact of inactive loss.}
Removing $\mathcal{L}_{\textrm{inact}}$ leads to loss of controllability with the edge hypernet predominantly predicting the full original anchor net, with minimal branching, leading to high resource usage and poor uniformity (Table~\ref{tab:nyu_ablation}).

\noindent \textbf{Impact of weighting factors.} 
Removing the two branching weights, $\frac{L-l}{L}$ and $A$, in the active loss, we make three key observations in Table~\ref{tab:nyu_ablation}: 1) average resource usage increases, 2) uniformity drops due to poor alignment between architectures and preferences, with larger architectures incorrectly predicted for skewed preferences, which ideally require less resources, 3) hypervolume remains almost constant across different $c$ indicating poor cost control. Resource usage plots are presented in the appendix.

\noindent \textbf{Analysis of task threshold.} 
We compare the effect of varying the threshold $\tau$ in Figure~\ref{fig:thresh}. Increasing the value beyond $1/N$ ($\sim 0.3$) leads to loss of controllability as indicated by the constant hypervolume across different values of $c$. This is due to the inability to account for uniform preferences. On the other hand, choosing values below this threshold leads to comparable performance. Additional explanations are provided in the appendix. 

\noindent \textbf{Task classification.}
We analyse the importance of the induced task dichotomy by considering all tasks as active. This leads to: 1) high overall resource usage, and 2) poor controllability, especially at low values of $c$, as shown in Table~\ref{tab:nyu_ablation}. Resource usage plots are presented in the appendix.
\begin{table}[]
\centering
\resizebox{\columnwidth}{!}{
\begin{tabular}{@{}lcccccc@{}}
\toprule
\multicolumn{1}{c}{\multirow{2}{*}{\textbf{Method}}} & \multicolumn{2}{c}{\textbf{HV.}$\uparrow$} & \multicolumn{2}{c}{\textbf{Unif.}$\uparrow$} & \multicolumn{2}{c}{\#\textbf{Inference Params}$\downarrow$} \\ \cmidrule(l){2-7}
\multicolumn{1}{c}{} & $c=0.0$ & $c=1.0$ & $c=0.0$ & $c=1.0$ & $c=0.0$ & $c=1.0$ \\ \midrule
Ours
&13.43      &13.08      &0.76       &0.78       &41.06M      &34.68M        \\ \midrule
no $\mathcal{L}_{\textrm{inactive}}$                    
&12.81      &12.73       &0.49       &0.51       &61.75M       &54.78M      \\
no layer weighting
&12.69       &12.21       &0.51       &0.53       &46.21M       &41.33M       \\
no task affinity $A$
&12.53       &12.35       &0.51       &0.52       &45.73M       &42.55M       \\ 
no task dichotomy 
&12.57      &11.20        &0.60       &0.63       &56.73M       &45.17M       \\\bottomrule
\end{tabular}}
\vspace{-0.25cm}
\caption{\textbf{Ablation study on NYU-v2}}
\vspace{-0.3cm}
\label{tab:nyu_ablation}
\end{table}
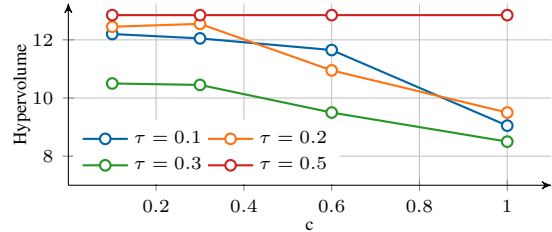
\begin{figure}[t]
\begin{tikzpicture}[>=stealth']
\begin{axis}[
    width=8cm,
    height=4cm,
    grid=major,
    major grid style={line width=.2pt,draw=gray!50},
    xmin=0,
    xmax=1.1,
    ymin=7.0,
    ymax=13.25,
    axis lines=middle,
    axis line style={->},
    x label style={at={(axis description cs:0.5,-0.15)},anchor=north},
    y label style={at={(axis description cs:-0.06,.5)},rotate=90,anchor=south},
    y tick label style={
        /pgf/number format/.cd,
        fixed,
        fixed zerofill,
        precision=0,
        /tikz/.cd
    },
    xlabel={c},
    ylabel={Hypervolume},
    legend style={at={(0.01, 0.005)}, anchor=south west, draw=none},
    legend cell align={left},
    legend columns=2,
]
\addplot[color=plotsblue,solid,thick,mark=*, mark options={fill=white,solid}] coordinates {
    (0.1, 12.2)
    (0.3, 12.05)
    (0.6, 11.65)
    (1.0, 9.05)
};
\addlegendentry{$\tau=0.1$};
\addplot[color=plotsorange,solid,thick,mark=*, mark options={fill=white,solid}] coordinates {
    (0.1, 12.45)
    (0.3, 12.55)
    (0.6, 10.95)
    (1.0, 9.50)
};
\addlegendentry{$\tau=0.2$};
\addplot[color=plotsgreen,solid,thick,mark=*, mark options={fill=white,solid}] coordinates {
    (0.1, 10.5)
    (0.3, 10.45)
    (0.6, 9.50)
    (1.0, 8.50)
};
\addlegendentry{$\tau=0.3$}
\addplot[color=plotsred,solid,thick,mark=*, mark options={fill=white,solid}] coordinates {
    (0.1, 12.85)
    (0.3, 12.85)
    (0.6, 12.85)
    (1.0, 12.85)
};
\addlegendentry{$\tau=0.5$}
\end{axis}
\end{tikzpicture} 
\vspace{-0.32cm}
\caption{\textbf{Varying thresholds on NYU-v2}}
\label{fig:thresh}
\vspace{-0.65cm}
\end{figure}

\section{Conclusion}
We present a new framework for dynamic resource allocation in multi-task networks. We design a controller using hypernets to dynamically predict both network architecture and weights to match user-defined task trade-offs and resource constraints. In contrast to current dynamic MTL methods which work with a fixed model, our formulation allows the flexibility in controlling the total compute cost and matches the task preference better. We show the effectiveness of our approach on four multi-task settings, attaining diverse and efficient architectures across a wide range of preferences.

\noindent\textbf{Limitations and future work.} Our framework searches solely over network width and thus, the compute cost is lower bounded by network depth.
One possible solution is to extend the search space to allow skip connections within and across streams to allow variable depth. 
Also, scalability could be an issue as the required memory for the anchor net is proportional to the number of tasks. Our future work will address these issues by reducing the dependency on the anchor net initialization.

\noindent\textbf{Acknowledgements.} This work was a part of Dripta S. Raychaudhuri’s internship at NEC Labs America. This work was also partially supported by the NRI grant 2021-67022-33453 and the NSF grant 1724341.


%

{\small
\bibliographystyle{ieee_fullname}
\bibliography{refs}
}
\appendix 
\section*{Appendix}
\section{Resource Usage Plots}
\begin{figure}[ht]
    \centering
    \includegraphics[width=0.8\columnwidth]{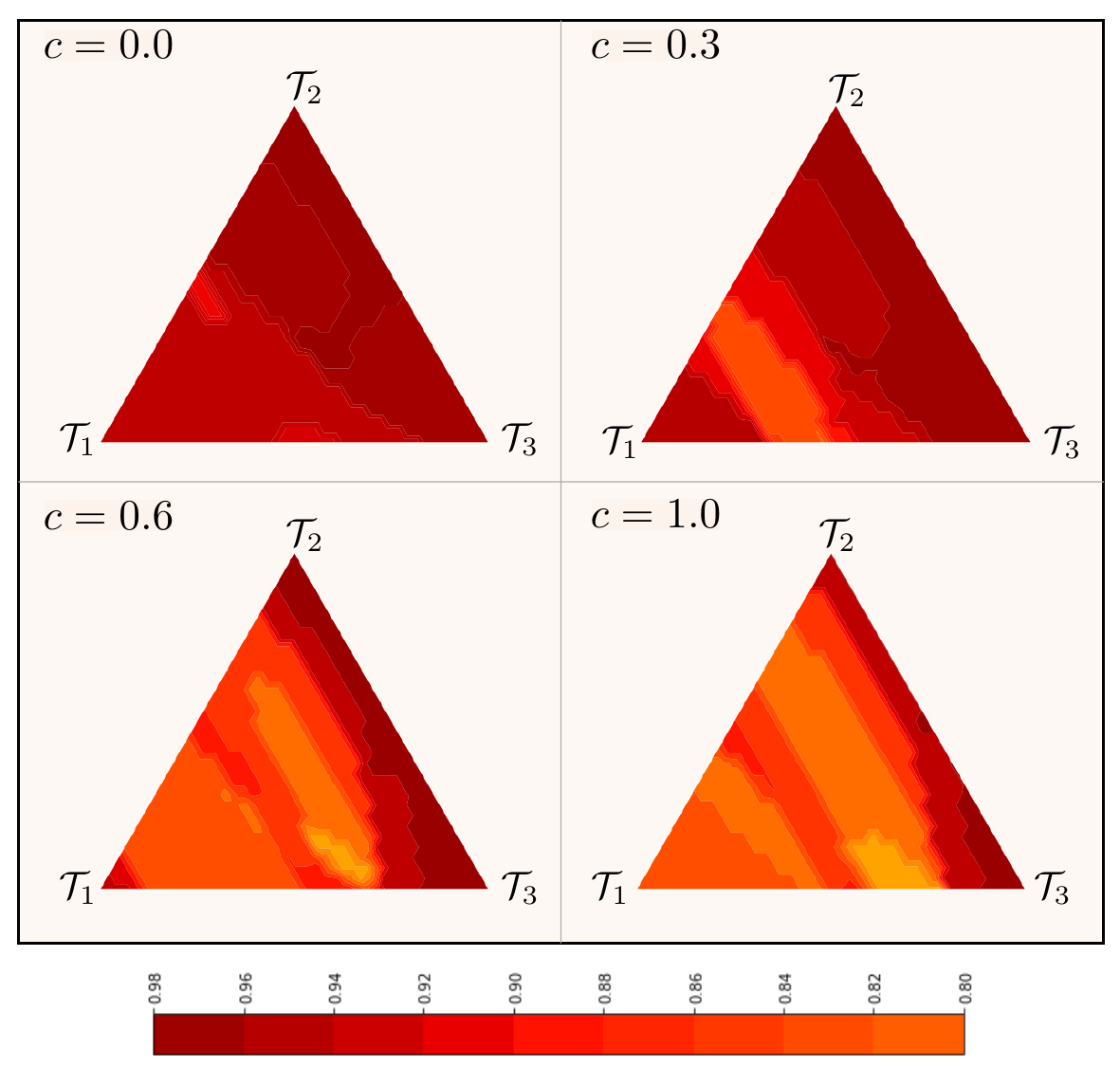}
    \vspace{-0.25cm}
    \caption{\textbf{Resource usage on removing $\mathcal{L}_{\textrm{inact}}$}}
    \label{fig:cost_no_inact}
    \vspace{-0.3cm}
\end{figure}

\begin{figure}[ht]
    \centering
    \includegraphics[width=0.8\columnwidth]{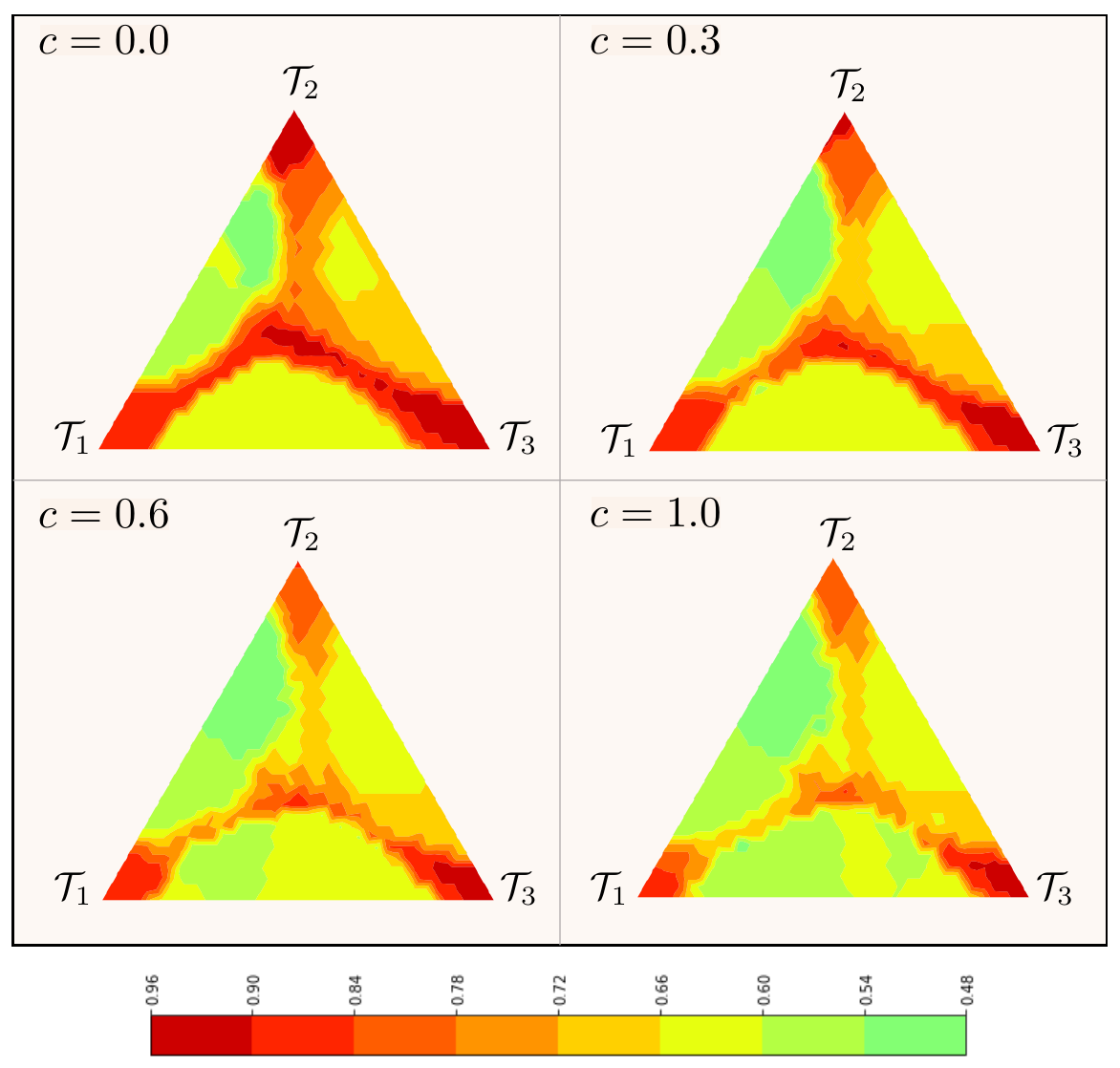}
    \vspace{-0.25cm}
    \caption{\textbf{Resource usage on removing $\frac{L-l}{L}$}}
    \label{fig:cost_no_w}
    \vspace{-0.3cm}
\end{figure}

\begin{figure}[ht]
    \centering
    \includegraphics[width=0.8\columnwidth]{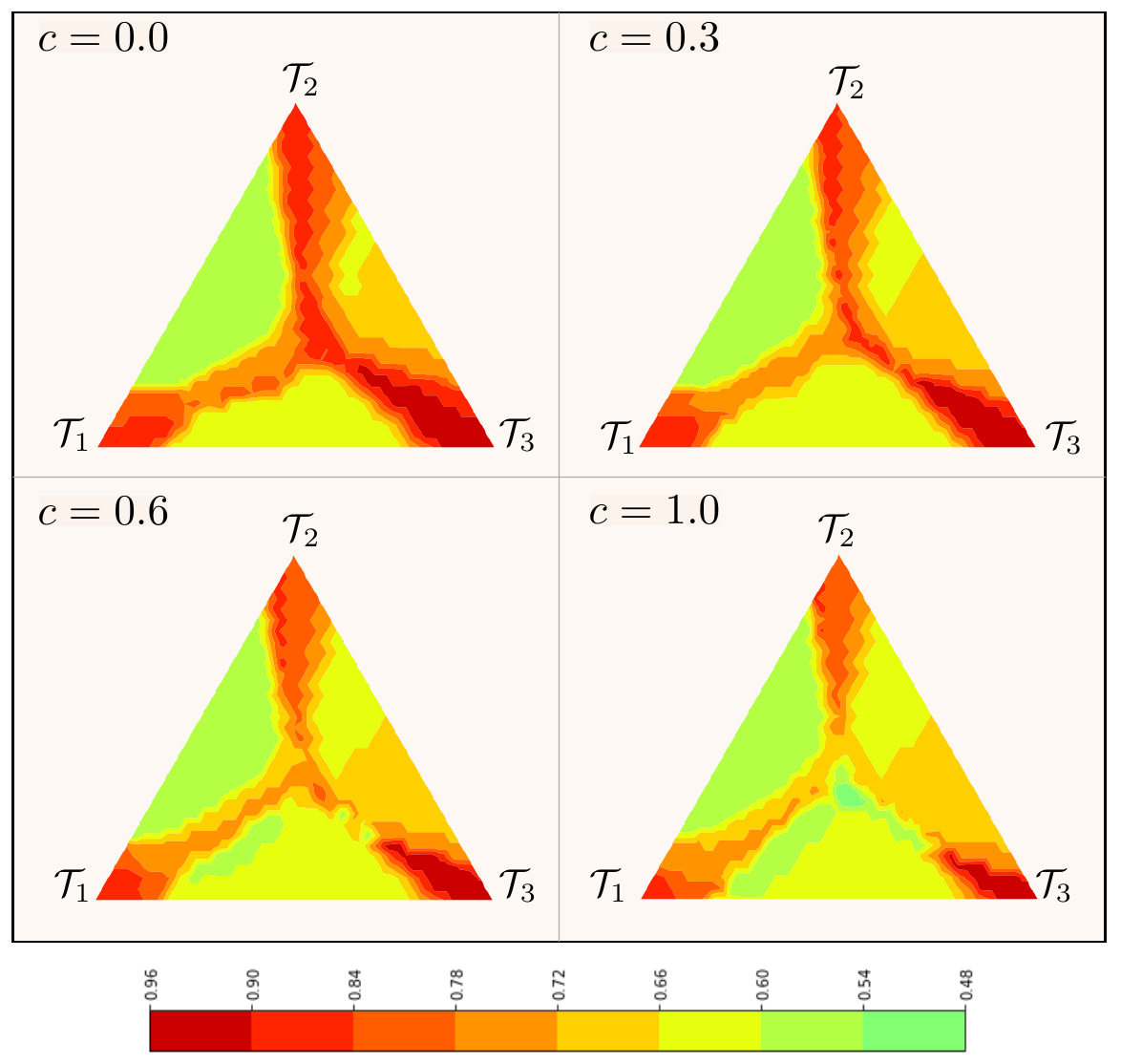}
    \vspace{-0.25cm}
    \caption{\textbf{Resource usage on removing $A$}}
    \label{fig:cost_no_corr}
    \vspace{-0.3cm}
\end{figure}

\begin{figure}[ht]
    \centering
    \includegraphics[width=0.8\columnwidth]{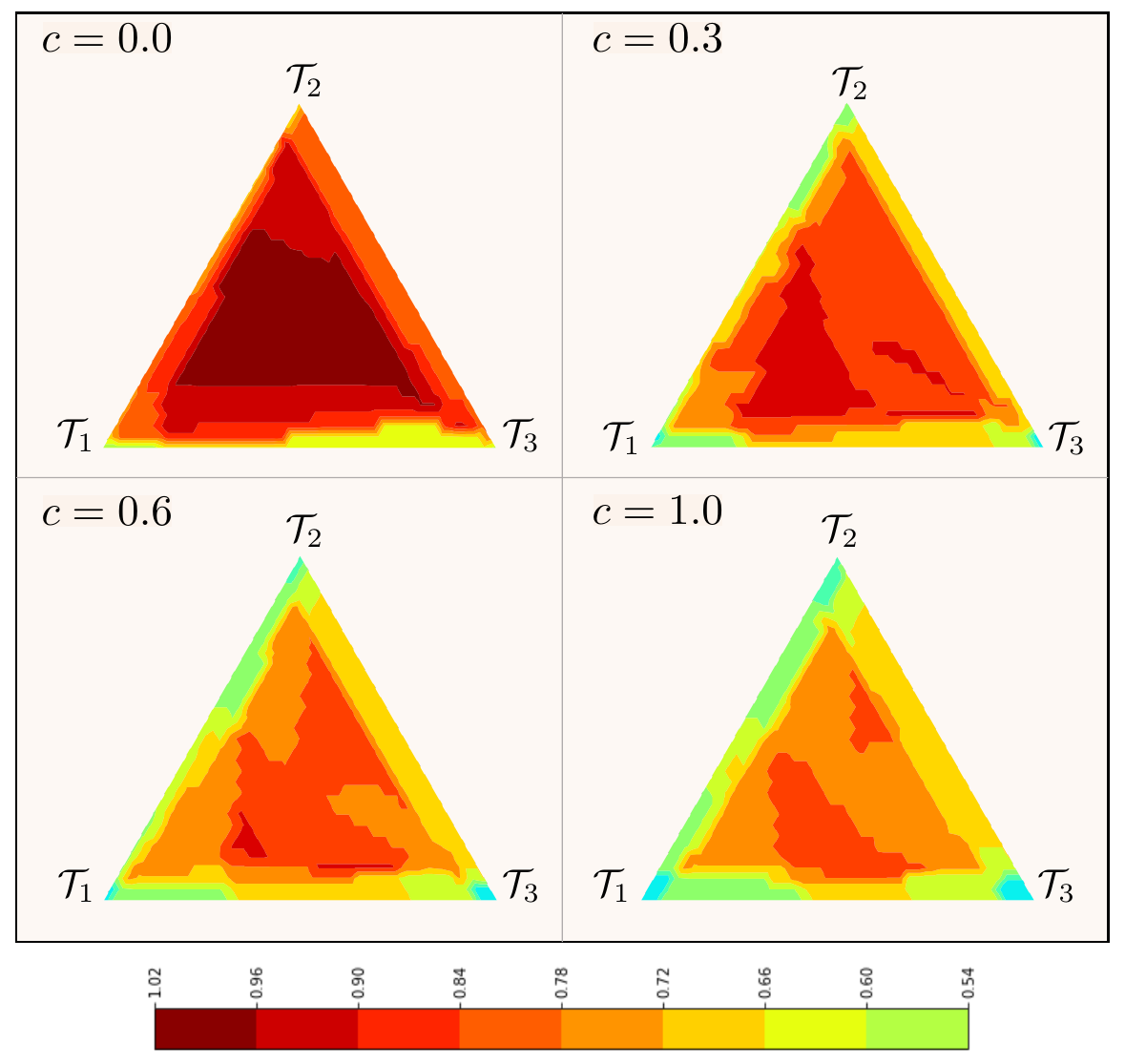}
    \vspace{-0.25cm}
    \caption{\textbf{Resource usage on removing task dichotomy}}
    \label{fig:cost_no_dichotomy}
    \vspace{-0.3cm}
\end{figure}

In Figures~\ref{fig:cost_no_inact}-\ref{fig:cost_no_dichotomy}, we plot the resource usage across different preferences on the NYU-v2 dataset to analyze the effect of different parts of our framework. The primary observation in each case is the lack of proper controllability. On removing $\mathcal{L}_{\textrm{inact}}$ or task dichotomy (Figures~\ref{fig:cost_no_inact},\ref{fig:cost_no_dichotomy}), the overall resource usage increases across all preferences. On the other hand, removing the active loss branching weights (Figures~\ref{fig:cost_no_w},\ref{fig:cost_no_corr}) leads to a misallocation of resources - more compute cost is allocated to skewed resources than the dense ones.


\begin{figure*}[ht]
\centering
\begin{subfigure}[b]{0.32\textwidth}
\includegraphics[width=\linewidth]{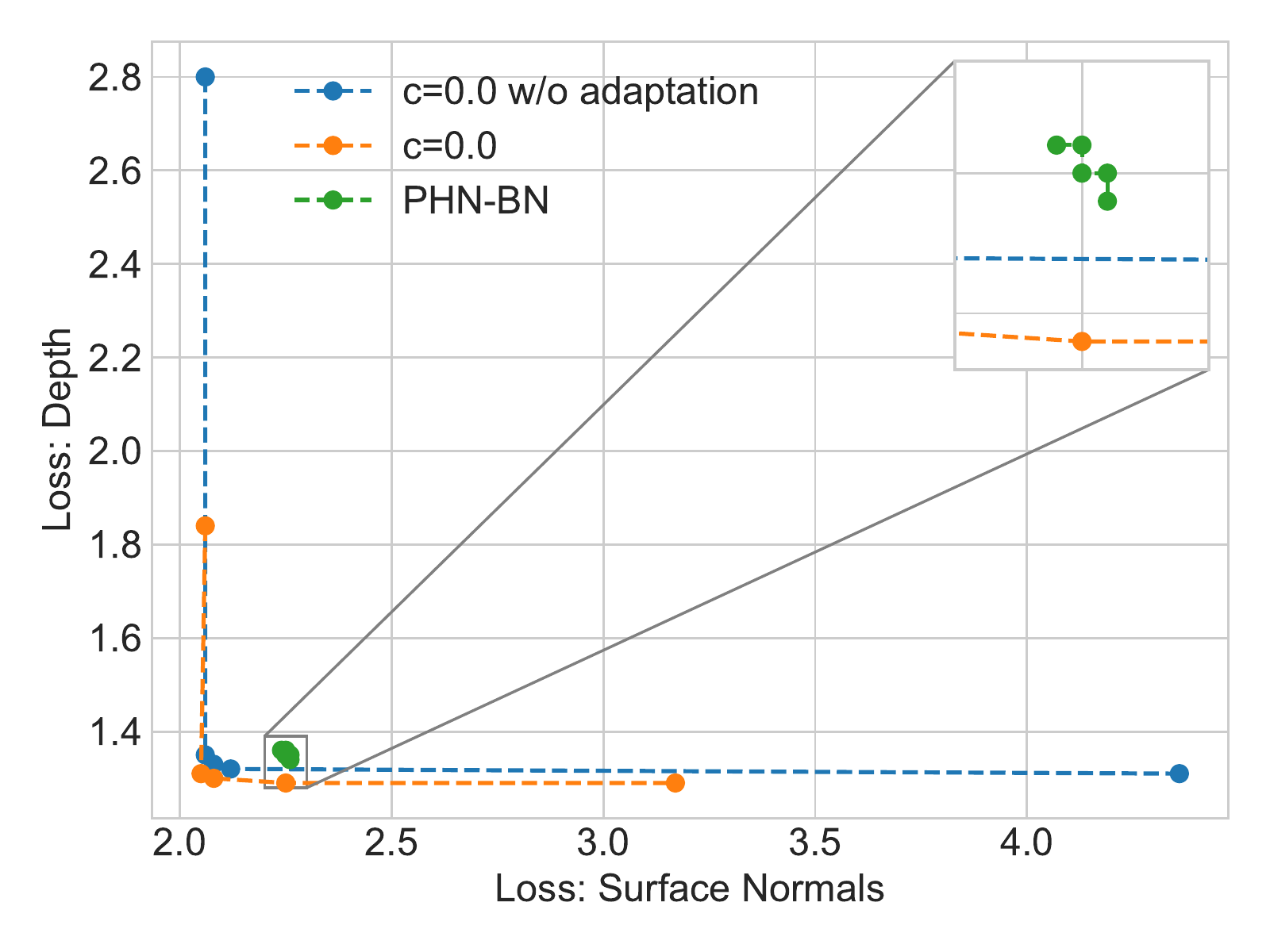}
\caption{Surface Normal/Depth}
\label{fig:nyu_pareto_normal_depth}
\end{subfigure}
\begin{subfigure}[b]{0.32\textwidth}
\includegraphics[width=\linewidth]{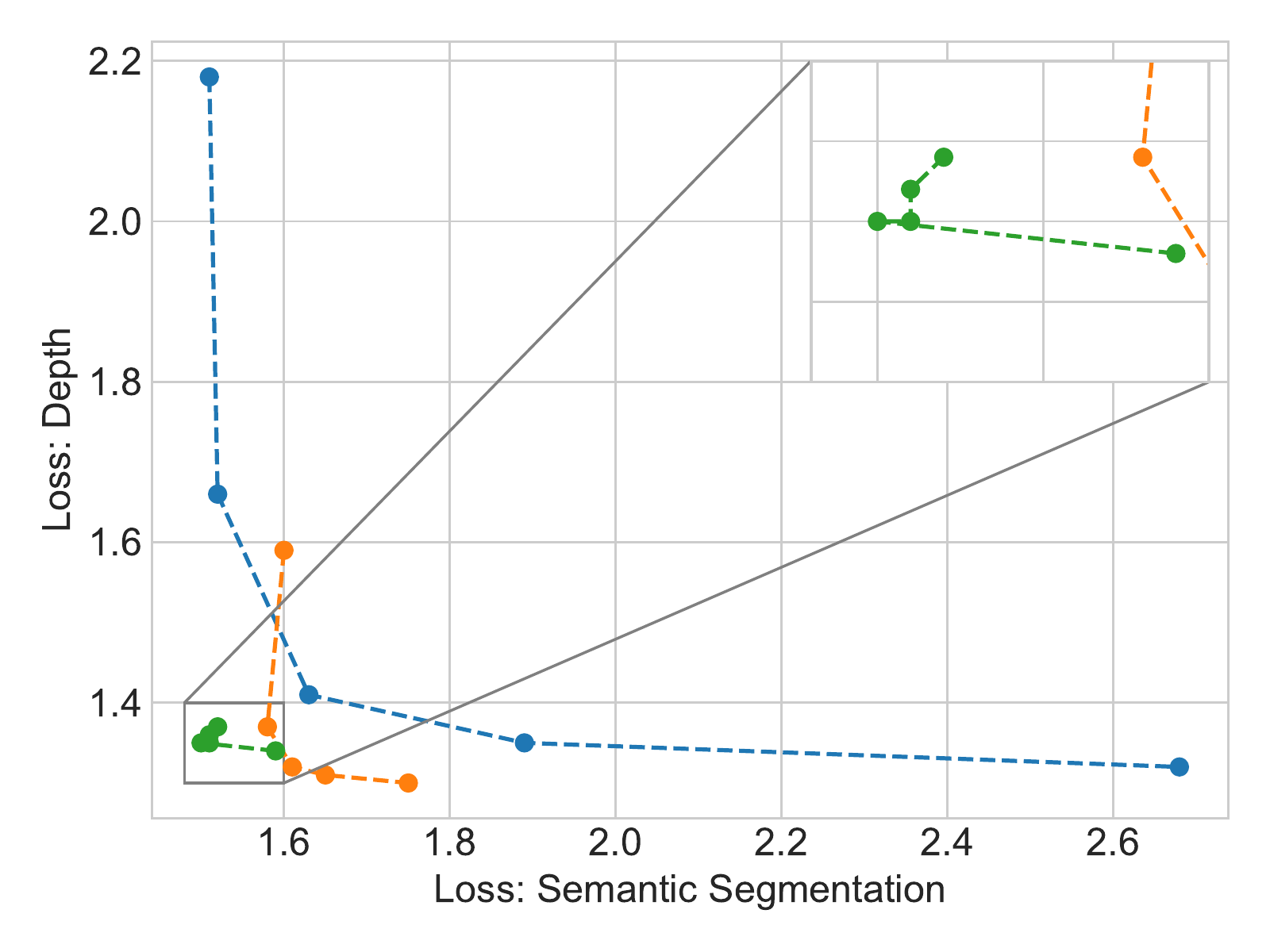}
\caption{Semantic Seg./Depth}
\label{fig:nyu_pareto_semseg_depth}
\end{subfigure}
\begin{subfigure}[b]{0.32\textwidth}
\includegraphics[width=\linewidth]{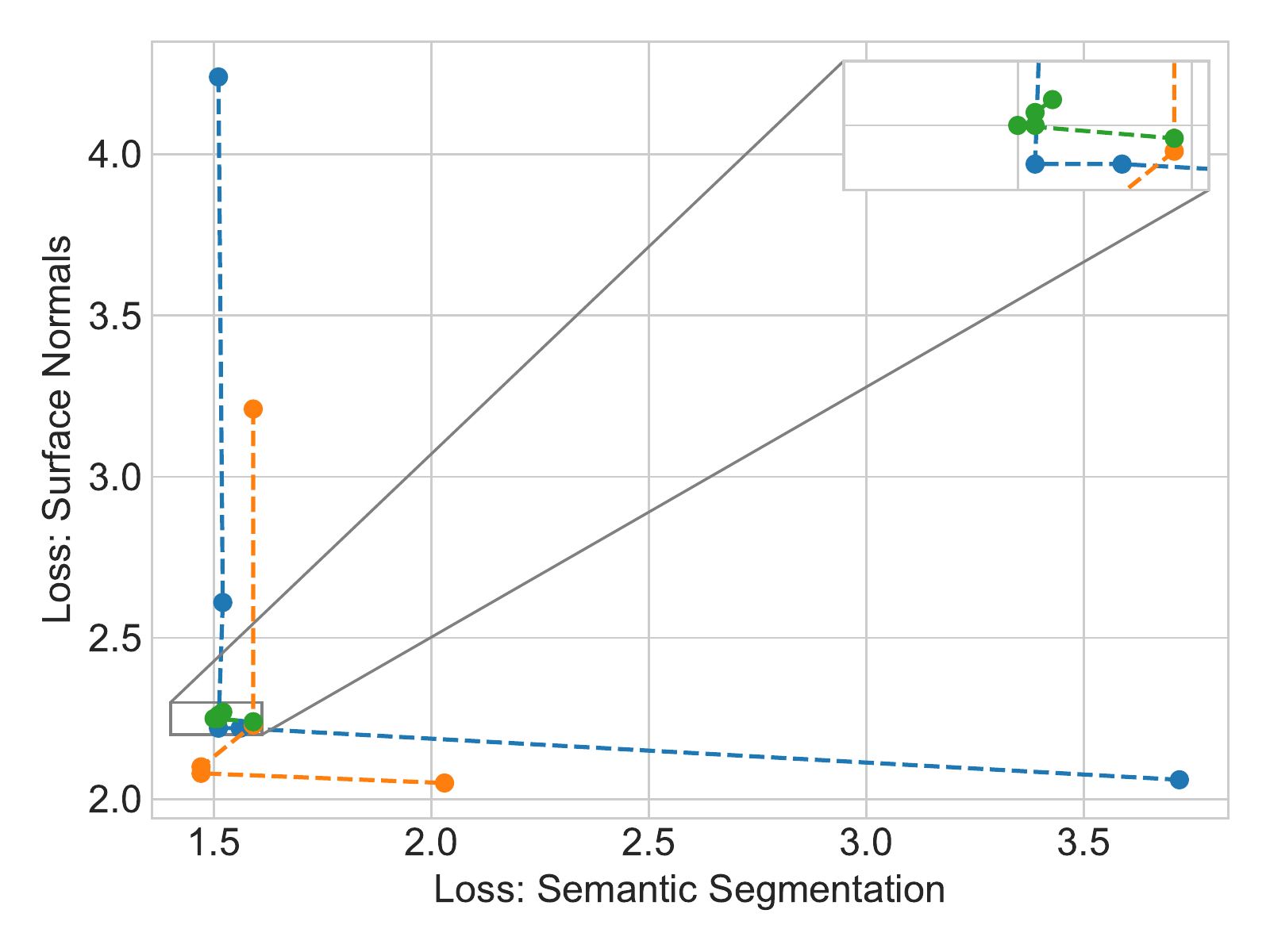}
\caption{Semantic Seg./Surface Normal}
\label{fig:nyu_pareto_semseg_normal}
\end{subfigure}
\vspace{-0.25cm}
\caption{\textbf{Trade-off curves on NYU-v2:} Visualization of performance trade-off between pairs of tasks.}
\vspace{-0.3cm}
\label{fig:pareto_nyu}
\end{figure*}

\begin{figure*}[ht]
\centering
\begin{subfigure}[b]{0.32\textwidth}
\includegraphics[width=\linewidth]{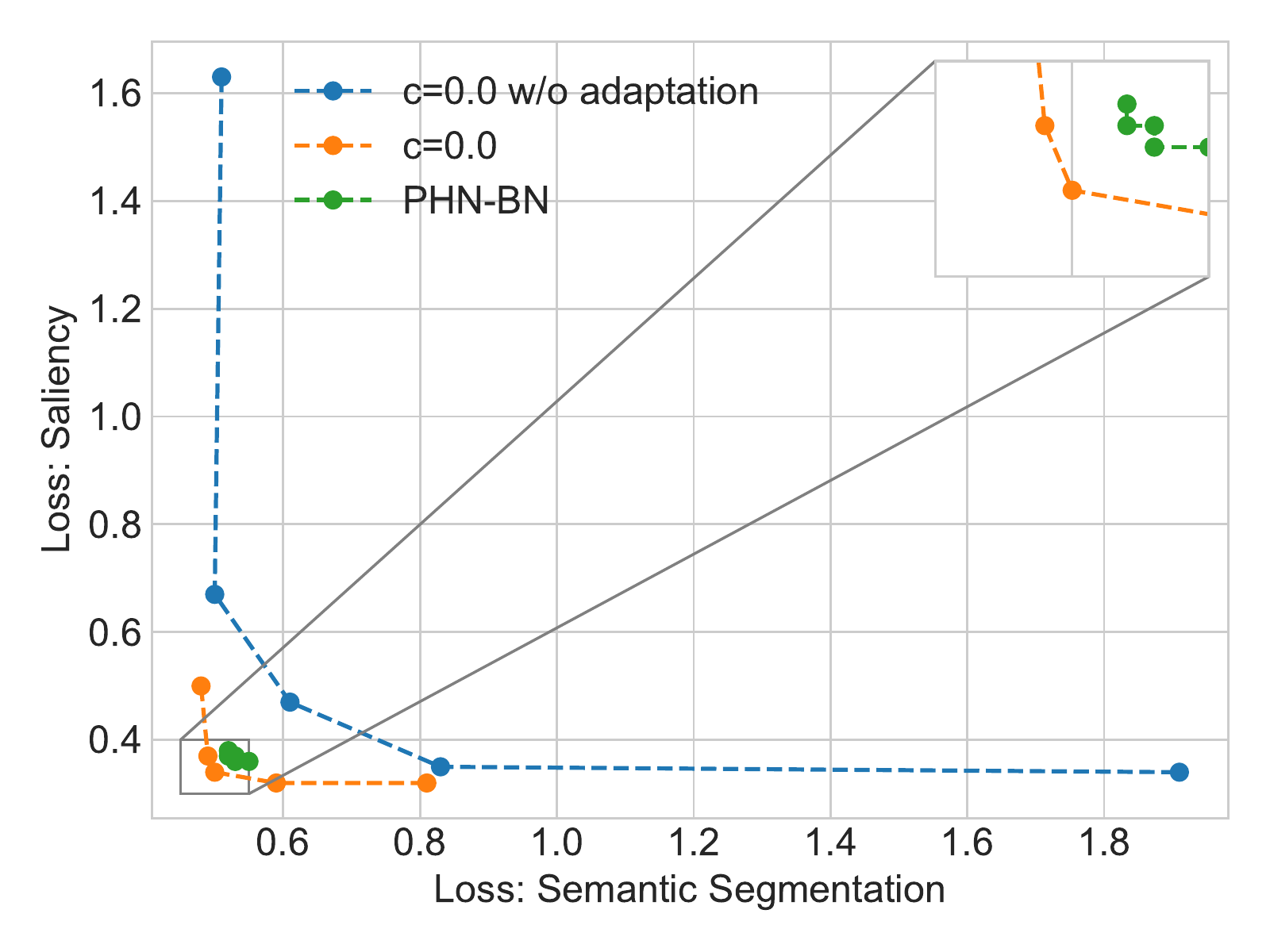}
\caption{Human Parts/Saliency}
\label{fig:pascal_pareto_parts_sal}
\end{subfigure}
\begin{subfigure}[b]{0.32\textwidth}
\includegraphics[width=\linewidth]{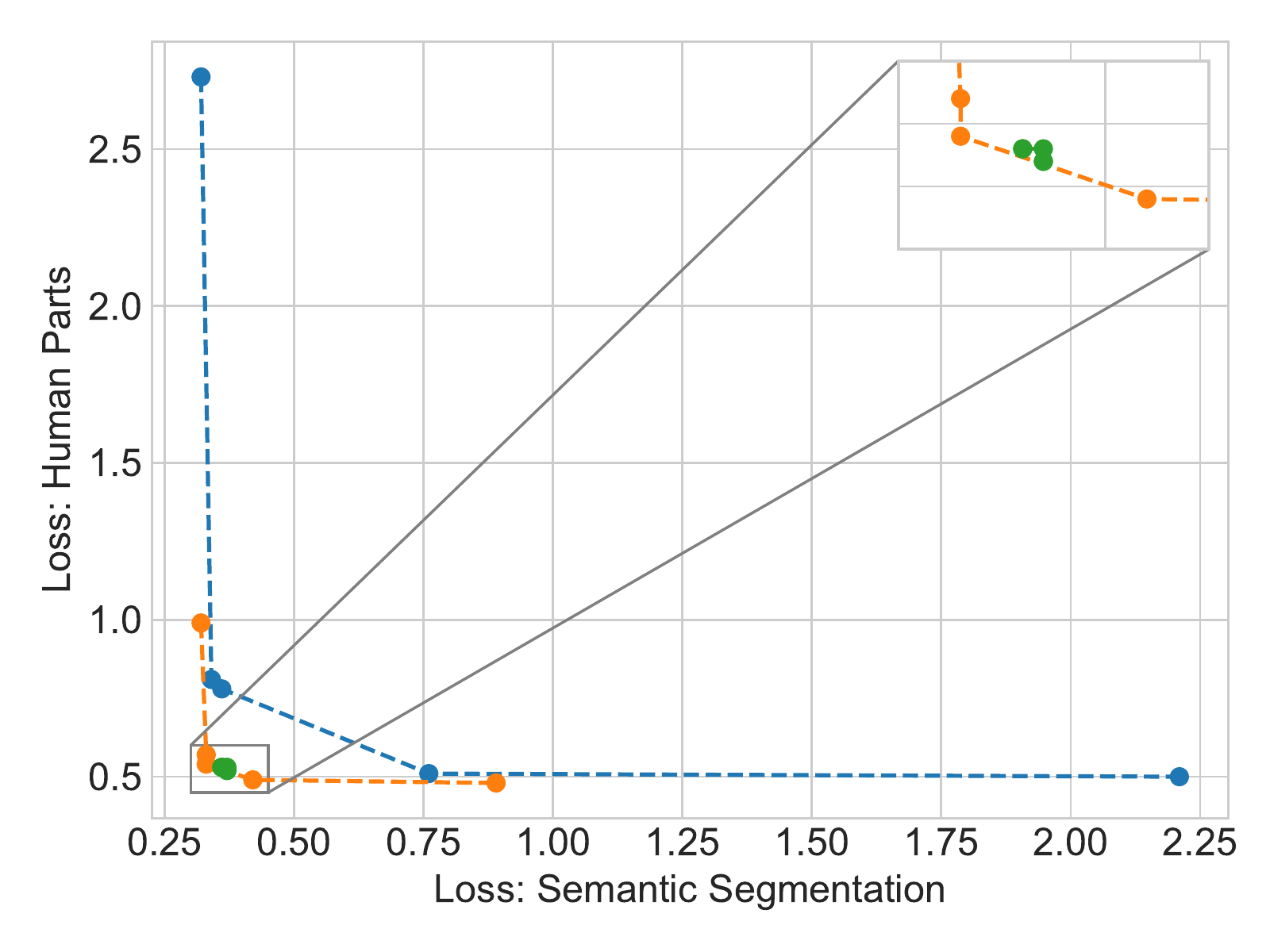}
\caption{Semantic Seg./Human Parts}
\label{fig:pascal_pareto_semseg_parts}
\end{subfigure}
\begin{subfigure}[b]{0.32\textwidth}
\includegraphics[width=\linewidth]{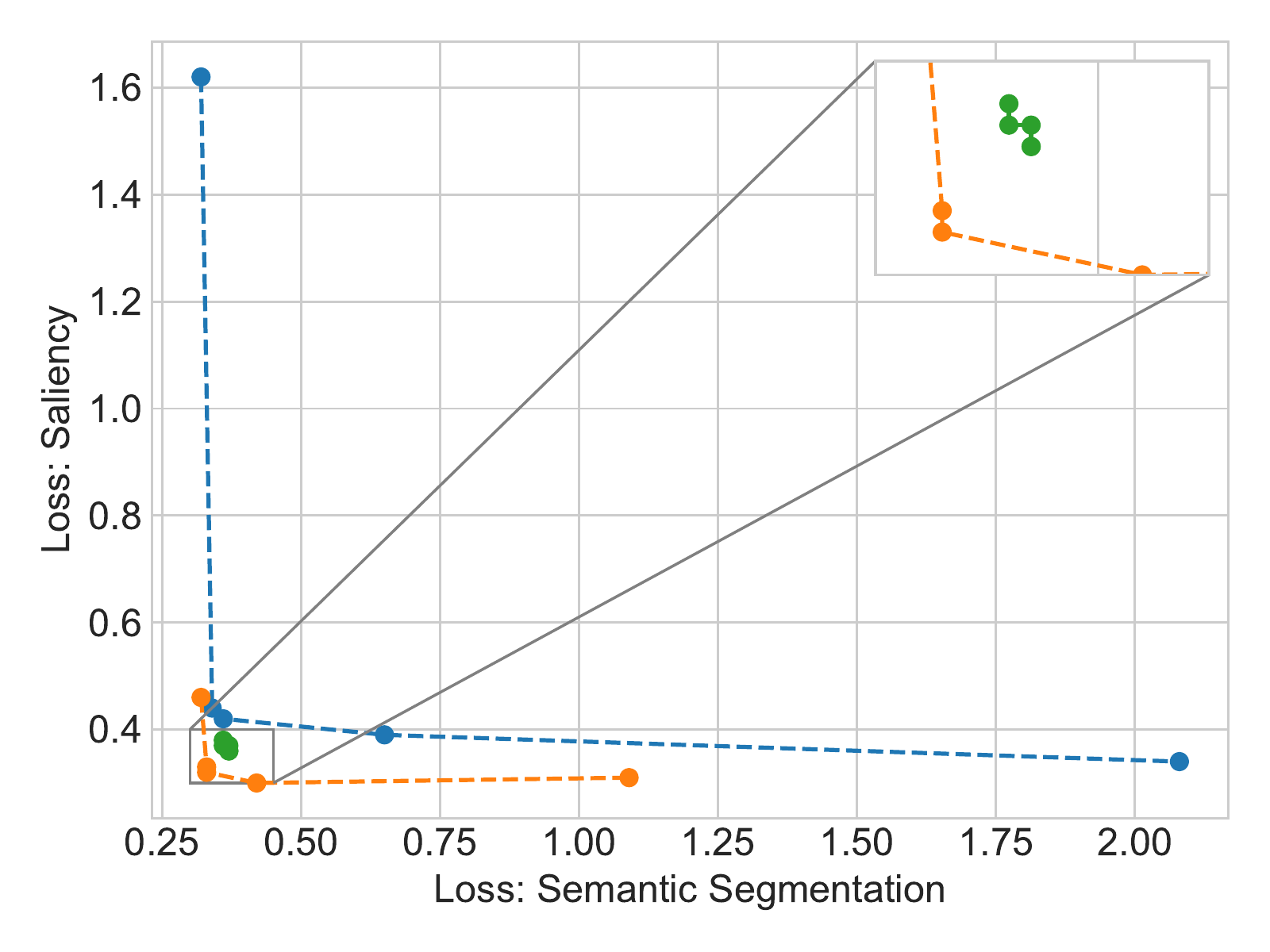}
\caption{Semantic Seg./Saliency}
\label{fig:pascal_pareto_semseg_sal}
\end{subfigure}
\vspace{-0.25cm}
\caption{\textbf{Trade-off curves on PASCAL-Context (3 tasks):} Visualization of performance trade-off between pairs of tasks.}
\vspace{-0.5cm}
\label{fig:pareto_pascal}
\end{figure*}

\begin{figure}\centering%
\begin{tikzpicture}[>=stealth',mark size=1.5pt]
\begin{axis}[
    width=4.5cm,
    height=3cm,
    grid=major,
    major grid style={line width=.2pt,draw=gray!50},
    xmin=0,
    xmax=1.0,
    ymin=0.25,
    ymax=1.5,
    axis lines=middle,
    axis line style={->},
    x label style={at={(axis description cs:0.5,-0.225)},anchor=north},
    y label style={at={(axis description cs:-0.175,.5)},rotate=90,anchor=south},
    y tick label style={
        /pgf/number format/.cd,
        fixed,
        fixed zerofill,
        precision=1,
        /tikz/.cd
    },
    xlabel={Task preference},
    ylabel={Loss},
    legend style={at={(0.45, 0.8)}, anchor=west, draw=none},
    legend cell align={left},
    title={Semantic Segmentation},
    every axis title/.style={anchor=south, at={(0.5,1.0)}},
]
\addplot[color=plotsorange,solid,thick,mark=*, mark options={fill=white,solid}] coordinates {
    (0.1, 1.44)
    (0.3, 0.56)
    (0.5, 0.36)
    (0.7, 0.34)
    (0.9, 0.32)
};
\addplot[color=plotsblue,solid,thick,mark=*, mark options={fill=white,solid}] coordinates {
    (0.1, 1.39)
    (0.3, 0.51)
    (0.5, 0.35)
    (0.7, 0.34)
    (0.9, 0.32)
};
\addplot[color=plotsorange,dashed,thick,mark=*, mark options={fill=white,solid}] coordinates {
    (0.1, 1.02)
    (0.3, 0.35)
    (0.5, 0.33)
    (0.7, 0.33)
    (0.9, 0.32)
};
\addplot[color=plotsblue,dashed,thick,mark=*, mark options={fill=white,solid}] coordinates {
    (0.1, 1.01)
    (0.3, 0.35)
    (0.5, 0.33)
    (0.7, 0.32)
    (0.9, 0.32)
};
\end{axis}
\end{tikzpicture} 
\hfill
\begin{tikzpicture}[>=stealth',mark size=1.5pt]
\begin{axis}[
    width=4.5cm,
    height=3cm,
    grid=major,
    major grid style={line width=.2pt,draw=gray!50},
    xmin=0,
    xmax=1.0,
    ymin=0.35,
    ymax=1.5,
    axis lines=middle,
    axis line style={->},
    x label style={at={(axis description cs:0.5,-0.225)},anchor=north},
    y label style={at={(axis description cs:-0.175,.5)},rotate=90,anchor=south},
    y tick label style={
        /pgf/number format/.cd,
        fixed,
        fixed zerofill,
        precision=1,
        /tikz/.cd
    },
    xlabel={Task preference},
    ylabel={Loss},
    legend style={at={(0.45, 0.8)}, anchor=west, draw=none},
    legend cell align={left},
    title={Human Parts},
    every axis title/.style={anchor=south, at={(0.5,1.0)}},
]
\addplot[color=plotsorange,solid,thick,mark=*, mark options={fill=white,solid}] coordinates {
    (0.1, 1.48)
    (0.3, 0.83)
    (0.5, 0.77)
    (0.7, 0.51)
    (0.9, 0.51)
};
\addplot[color=plotsblue,solid,thick,mark=*, mark options={fill=white,solid}] coordinates {
    (0.1, 1.47)
    (0.3, 0.83)
    (0.5, 0.73)
    (0.7, 0.51)
    (0.9, 0.50)
};
\addplot[color=plotsorange,dashed,thick,mark=*, mark options={fill=white,solid}] coordinates {
    (0.1, 0.81)
    (0.3, 0.58)
    (0.5, 0.53)
    (0.7, 0.49)
    (0.9, 0.49)
};
\addplot[color=plotsblue,dashed,thick,mark=*, mark options={fill=white,solid}] coordinates {
    (0.1, 0.81)
    (0.3, 0.57)
    (0.5, 0.53)
    (0.7, 0.49)
    (0.9, 0.47)
};
\end{axis}
\end{tikzpicture} 
\hfill
\begin{tikzpicture}[>=stealth',mark size=1.5pt]
\begin{axis}[
    width=4.5cm,
    height=3cm,
    grid=major,
    major grid style={line width=.2pt,draw=gray!50},
    xmin=0,
    xmax=1.0,
    ymin=0.2,
    ymax=1.75,
    axis lines=middle,
    axis line style={->},
    x label style={at={(axis description cs:0.5,-0.225)},anchor=north},
    y label style={at={(axis description cs:-0.175,.5)},rotate=90,anchor=south},
    y tick label style={
        /pgf/number format/.cd,
        fixed,
        fixed zerofill,
        precision=1,
        /tikz/.cd
    },
    xlabel={Task preference},
    ylabel={Loss},
    title={Saliency},
    every axis title/.style={anchor=south, at={(0.5,1.0)}},
    legend style={
        anchor=north west,
        at={(1.25,1.0)},
        text=black,
        draw=none,
        /tikz/every even column/.append style={column sep=0.2cm},
    },
    legend cell align={left},
]
\addplot[color=plotsblue,solid,thick,mark=*, mark options={fill=white,solid}] coordinates {
    (0.1, 1.62)
    (0.3, 0.55)
    (0.5, 0.42)
    (0.7, 0.35)
    (0.9, 0.34)
};
\addlegendentry{$c=0.0$ w/o adaptation};
\addplot[color=plotsorange,solid,thick,mark=*, mark options={fill=white,solid}] coordinates {
    (0.1, 1.62)
    (0.3, 0.88)
    (0.5, 0.41)
    (0.7, 0.35)
    (0.9, 0.34)
};
\addlegendentry{$c=1.0$ w/o adaptation};
\addplot[color=plotsblue,dashed,thick,mark=*, mark options={fill=white,solid}] coordinates {
    (0.1, 0.48)
    (0.3, 0.33)
    (0.5, 0.32)
    (0.7, 0.31)
    (0.9, 0.31)
};
\addlegendentry{$c=0.0$};
\addplot[color=plotsorange,dashed,thick,mark=*, mark options={fill=white,solid}] coordinates {
    (0.1, 0.48)
    (0.3, 0.36)
    (0.5, 0.32)
    (0.7, 0.31)
    (0.9, 0.31)
};
\addlegendentry{$c=1.0$};
\end{axis}
\end{tikzpicture} 
\vspace{-0.25cm}
\caption{\textbf{Marginal evaluation on PASCAL-Context (3 tasks)}}
\label{fig:marginal_pascal3}
\vspace{-0.3cm}
\end{figure}
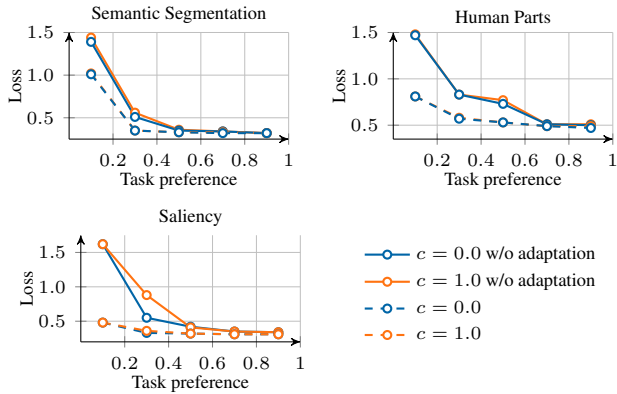
\section{Additional Results on Controllability}

\noindent\textbf{Trade-off curves.}  
In Figures~\ref{fig:pareto_nyu} and \ref{fig:pareto_pascal}, we visualize the task controllability on pairs of tasks while keeping the preference on the rest fixed to zero. Compared to conventional dynamic models, our framework achieves a much larger dynamic range in terms of performance.
\vskip 2pt
\noindent\textbf{Marginal evaluation.} 
We present the marginal evaluation of task controllability on the PASCAL-Context 3 task learning setting in Figure~\ref{fig:marginal_pascal3}.

\begin{table}[]
\centering
\resizebox{\columnwidth}{!}{
\begin{tabular}{@{}lccccccc@{}}
\toprule
\textbf{Method} & $\mathcal{T}_1$ $\uparrow$ & $\mathcal{T}_2$ $\uparrow$ & $\mathcal{T}_3$ $\uparrow$ & $\mathcal{T}_4$ $\downarrow$ & $\mathcal{T}_5$ $\downarrow$ & Avg $\Delta_{\mathbf{T}} (\%) \uparrow$ & \# \textbf{Params} (\%) $\downarrow$                  \\ \midrule
Single-Task                
& 63.88   
& 58.08  
& 64.92  
& 13.95    
& 0.018
& -
& -
\\ \midrule
LTB                        
& 59.45   
& 56.48
& 65.29
& 14.19
& 0.018
& -3.61
& -36.3 
\\
BMTAS    
& 63.27   
& 59.32
& 64.52
& 13.87
& 0.018
& +0.37
& -53.8    
\\ \midrule
Ours, c=0.0                
& 61.64   
& 56.95  
& 64.55  
& 13.72    
& 0.018
& -1.45
& -48.5  
\\ 
$\text{Ours}^\dagger$, c=0.0                
& 60.03 
& 56.91  
& 64.59  
& 13.95    
& 0.018
& -2.84
& -48.5
\\
Ours, c=1.0 
& 61.15   
& 56.85  
& 64.67  
& 13.74    
& 0.018
& -1.76
& -50.1
\\
$\text{Ours}^\dagger$, c=1.0  
& 59.74   
& 56.66  
& 64.62  
& 13.98    
& 0.018
& -3.20
& -50.1
\\ \bottomrule
\end{tabular}}
\vspace{-0.15cm}
\caption{\textbf{Architecture evaluation on PASCAL-Context (5 tasks).} We report the mean intersection over union for $\mathcal{T}_1:\text{Semantic seg.}$, $\mathcal{T}_2:\text{Parts seg.}$, and $\mathcal{T}_3:\text{Saliency}$. We report mean error in angle for $\mathcal{T}_4:\text{Surface normal}$ and mean loss for $\mathcal{T}_5:\text{Edge}$. Presence of $\dagger$ indicates that we train the networks from ImageNet weights, while its absence indicates training from anchor net weights.} \label{tab:pascal5_arch}
\vspace{-0.3cm}
\end{table}
\begin{figure*}[ht]
    \centering
    \includegraphics[width=0.93\textwidth]{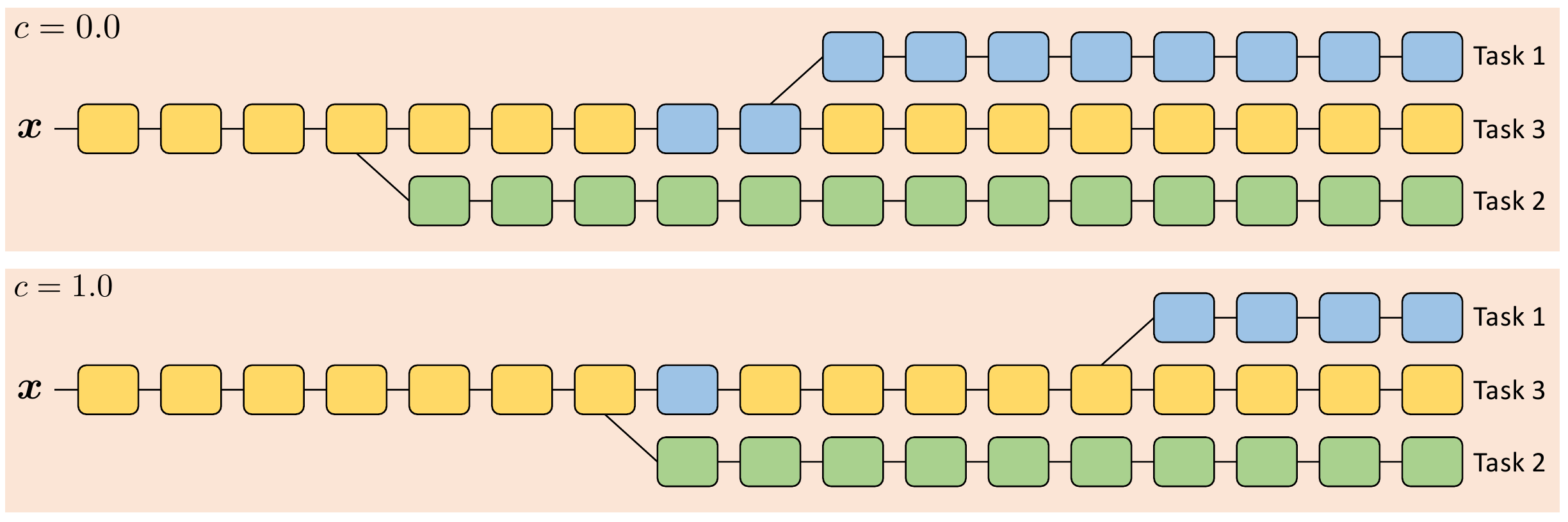}
    \vspace{-0.25cm}
    \caption{\textbf{Predicted architectures on NYU-v2 for uniform task preference.}}
    \label{fig:dense_arch}
    \vspace{-0.3cm}
\end{figure*}
\begin{figure*}[ht]
    \centering
    \includegraphics[width=0.93\textwidth]{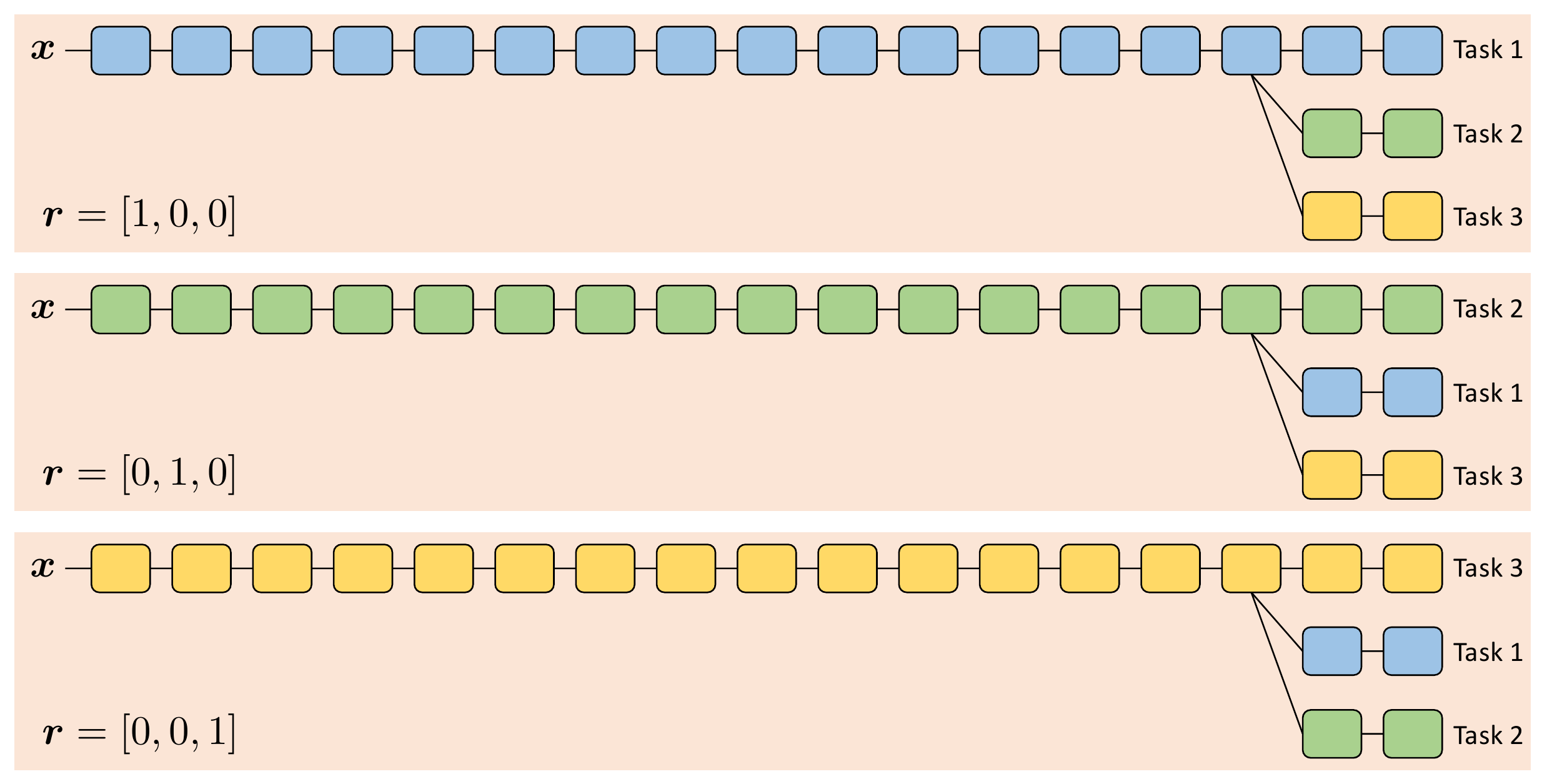}
    \vspace{-0.25cm}
    \caption{\textbf{Predicted architectures on NYU-v2 for preferences focusing on a single task.}}
    \label{fig:skewed_arch}
    \vspace{-0.5cm}
\end{figure*}

\section{Architecture Evaluation}
Figure~\ref{fig:dense_arch} illustrates the architectures predicted by the edge hypernet for uniform task preference on the NYU-v2 3-task setting. As we increase $c$, the model size decreases via increased sharing of the low-level features. We also visualize the architectures for skewed preferences in Figure~\ref{fig:skewed_arch}. This leads to architectures which are predominantly a single-stream network, with the selected stream corresponding to the primary task. In all cases, Task 1 denotes semantic segmentation, Task 2 denotes surface normals estimation, and Task 3 denotes depth estimation.
We report an additional evaluation of the predicted architectures on the Pascal-Context dataset in Table~\ref{tab:pascal5_arch}. We choose the architecture predicted for a uniform task preference and, similar to LTB, we retrain it for a fair comparison.

\section{Calculation of Task Affinity}
\label{sec:task_affinity}
We adopt \emph{Representational Similarity Analysis} (RSA)~\cite{dwivedi2019representation} to obtain the task affinity scores from the anchor net $\mathrm{F}$.
First, using a random subset of $K$ instances from the training set, we extract the features for each of these data points for every task. Let us denote the extracted feature map for instance $k$ at layer $l$ for task $i$ as $\bm{f}_{[i,l,k]}$. Using these feature maps, we compute the feature similarity tensor $\mathbf{S}_l$ at each layer, of dimensions $N\times K \times K$, as follows,
\begin{equation}
    \mathbf{S}_l \left(i,k,k'\right) = \frac{\langle\bm{f}_{[i,l,k]},\bm{f}_{[i,l,k']}\rangle}{\|\bm{f}_{[i,l,k]}\|\|\bm{f}_{[i,l,k']}\|}.
\end{equation}
The task dissimilarity matrix $D_l$ at layer $l$, of dimensions $N \times N$, is calculated as follows,
\begin{equation}
    D_l(i,j) = \|\mathbf{S}_l \left(i,:,:\right)-\mathbf{S}_l \left(j,:,:\right)\|_{F}.
\end{equation}
\begin{figure}[ht]
    \centering
    \includegraphics[width=0.9\columnwidth]{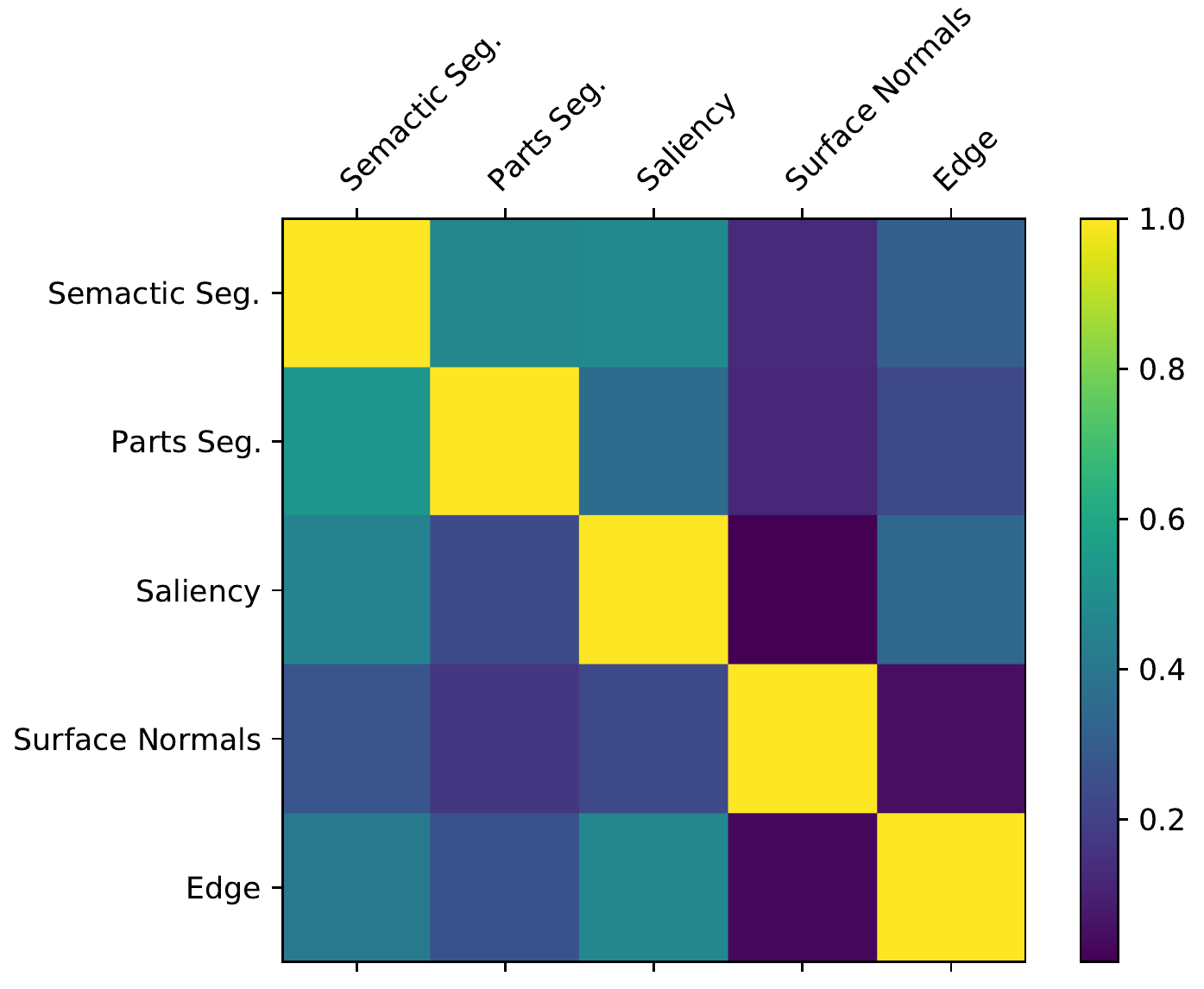}
    \vspace{-0.25cm}
    \caption{\textbf{Task affinity for PASCAL-Context 5-task setting}}
    \vspace{-0.5cm}
    \label{fig:task_rel}
\end{figure}
The rows of $D_l$ are normalized separately between $[0,1]$ to get the scaled task dissimilarity matrix $\hat{D}_l$. The task affinity matrix is subsequently calculated as $A_l(i,j)=1-\hat{D}_l(i,j).$ The factor $A(i,j)$ is obtained by taking the mean of the affinity scores across layers, i.e., $A(i,j)=\frac{1}{L}\sum_l A_l(i,j)$. Figure~\ref{fig:task_rel} presents an example of $A$ for the PASCAL-Context dataset. 

\section{Calculation of $P_{use}$}
$P_{use}(l,i)$ represents the probability that the node $i$ at layer $l$ is present in the tree structure that is sampled by the edge hypernet. We follow a dynamic programming approach to calculate this value. Denoting the $i^{th}$ node in the $l^{th}$ layer as $(l,i)$, we can define the marginal probability that the node $(l+1,k)$ samples the node $(l,i)$ as its parent as $\nu^{l}_k(i)$.  Thus, the probability that the node $(l,i)$ is used in the sampled tree structure is
\begin{equation} \label{eq:p_use}
    P_{use}(l,i)
    = 1 - \prod_{k}\{ 1 - P_{use}(l+1,k) \cdot \nu^{l}_k(i) \},
\end{equation}
where $P_{use}(L,i) = 1$ for all $i \in [N]$.

Note that the actual path chosen from hypernet can be different from the ones whose $\alpha$ is penalized in (3). We also tested an alternative, which applies dynamic programming similarly as in (\ref{eq:p_use}), to penalize exact activated path. Since the performance gain was minor, we adopt (3) for simplicity.

\section{Pseudo Code}
\begin{algorithm}[h]
\caption{Training controllable dynamic multi-task architecture.}
\begin{algorithmic}[1]

\State Initialize $\mathrm{F}$ with single-task weights \Comment{Anchor net}

\While{$h$ not converged} \Comment{Edge hypernet}
\State $\bm{r} \sim \textrm{Dir}(\eta), c \sim \textrm{Unif}(0,1)$
\State Sample batch $\{(\bm{x},\bm{y}_1,\dots,\bm{y}_N)\}$
\State $\hat{\bm{\alpha}}=h(\bm{r},c; \phi)$
\State Calculate $\Omega(\bm{r}, c, \hat{\bm{\alpha}})$ 
\State Calculate $\mathcal{L}_{\textrm{task}}(\bm{r}) = \sum_ir_i\mathcal{L}_i(\bm{x},\bm{y}_i,\mathrm{F},\hat{\bm{\alpha}})$
\State Update $\phi$ by back-propagating $\mathcal{L}_{\textrm{task}}+\Omega$
\EndWhile

\While{$\bar{h}$ not converged} \Comment{Weight hypernet}
\State $\bm{r} \sim \textrm{Dir}(\eta), c \sim \textrm{Unif}(0,1)$
\State Sample batch $\{(\bm{x},\bm{y}_1,\dots,\bm{y}_N)\}$
\State $\hat{\bm{\alpha}}=h(\bm{r},c; \phi)$
\State $\hat{\bm{\beta}},\hat{\bm{\gamma}}=\bar{h}(\bm{r},c; \bar{\phi})$
\State Calculate $\mathcal{L}_{\textrm{task}}(\bm{r}) = \sum_ir_i\mathcal{L}_i(\bm{x},\bm{y}_i,\mathrm{F},\hat{\bm{\alpha}},\hat{\bm{\beta}},\hat{\bm{\gamma}})$
\State Update $\bar{\phi}$ by back-propagating $\mathcal{L}_{\textrm{task}}$
\EndWhile

\end{algorithmic}
\label{alg:train}
\end{algorithm}

\section{Implementation Details} \label{sec:implementation_details}
\noindent \textbf{Hypernet architecture.} 
The edge hypernet is constructed using an MLP with two hidden layers (dimension 100) and $L$ linear heads, (dimension $N\times N$) which output the flattened branching distribution parameters at each layer. For the weight hypernet, we use a similar MLP with three hidden layers (dimension 100) and generate the normalization parameters using linear heads. In both cases we use learnable embeddings for each task ($e_i$) and the cost ($e_c$). Given the user preference $(\bm{r},c)$, the preference embedding is calculated as $p=\sum_i r_ie_i + ce_c$. This embedding $p$ is then used as input the MLP. The preference dimension is set to 32 in all experiments. PHN-BN and PHN~\cite{navon2021learning} use a similar architecture to the weight hypernet, with PHN employing an additional chunking~\cite{ha2016hypernetworks} embedding for scalability.
\vskip 2pt
\noindent \textbf{Task loss scaling.} Due to the different scales of the various task losses, we first weight the loss terms with a factor $w_i$ before applying linear scalarization with respect to $\bm{r}$. This ensures that the relative task importance is not skewed by the different loss scales. Thus, $\mathcal{L}_{\textrm{task}}(\bm{r})=\sum_ir_iw_i\mathcal{L}_i$.
\vskip 2pt
\noindent \textbf{Anchor net architecture.} The anchor net comprises $N$ single-task networks, with each stream corresponding to a particular task. For experiments on dense prediction tasks, we use the DeepLabv3+ architecture~\cite{chen2018encoder} for each task. The MobileNetV2~\cite{sandler2018mobilenetv2} backbone is used for experiments on PASCAL-Context~\cite{mottaghi2014the}, while the ResNet-34~\cite{he2016deep} backbone is used for experiments on NYU-v2~\cite{silberman2012indoor}. For CIFAR-100~\cite{krizhevsky2009learning} experiments we use the ResNet-9~\footnote{\url{https://github.com/davidcpage/cifar10-fast}} architecture with linear task heads.
\vskip 2pt
\noindent \textbf{Hyperparameters.}
For experiments on CIFAR-100 we use $\lambda_\mathcal{A}=0.2$ and $\lambda_\mathcal{I}=0.02$. For the rest, we use $\lambda_\mathcal{A}=1$ and $\lambda_\mathcal{I}=0.1$. The task scaling weights are given below:
\begin{enumerate}[noitemsep,nolistsep]
    \item \textbf{NYU-v2}: 
    \begin{itemize}[noitemsep,nolistsep]
        \item Semantic segmentation: 1
        \item Surface normals: 10
        \item Depth: 3
    \end{itemize}
    \item \textbf{PASCAL-Context}: 
    \begin{itemize}[noitemsep,nolistsep]
        \item Semantic segmentation: 1
        \item Human parts segmentation: 2
        \item Saliency: 5
        \item Surface normals: 10
        \item Edge: 50
    \end{itemize}
\end{enumerate}
For CIFAR-100, all tasks are equally weighted. 

The threshold $\tau$ is set to $0.02$ for CIFAR-100, $0.2$ for NYU-v2 and PASCAL-Context (3 task), and $0.1$ for PASCAL-Context (5 task). 
\vskip 2pt
\noindent \textbf{Training.} Hypernetworks are trained using Adam for 30K steps with a learning rate of 1$e-$3, reduced by a factor of 0.3 every 14K steps. Temperature $\zeta$ is initialized to 5 and is decayed by 0.97 every 300 steps. Single-task networks for dense-prediction tasks are trained in accordance to~\cite{bruggemann2020automated}. For CIFAR, we use Adam with a learning rate of 1$e-$3 and weight decay of 1$e-$5 for 75 epochs.
\vskip 2pt
\begin{table}[t] 
\centering
\resizebox{0.6\columnwidth}{!}{%
\begin{tabular}{lllll}
\toprule
$\eta$   & 0.1 & 0.2 & 0.5 & 1.0 \\ \midrule
c=0.0 &4.20     &4.26     &4.26     &4.26     \\
c=1.0 &4.18     &4.25     &4.25     &4.21     \\ \bottomrule
\end{tabular}%
}
\caption{\textbf{Varying parameterization of Dirichlet distribution.}}
\vspace{-0.3cm}
\label{table:pref}
\end{table}
\noindent \textbf{Preference sampling.}
During training we sample preferences $(\bm{r},c)$ from the distribution $P_{(\bm{r},c)}=P_{\bm{r}}P_c$, where $P_{\bm{r}}$ is defined as a Dirichlet distribution of order $N$ with parameter $\eta=[\eta_1,\eta_2,\dots,\eta_N ]$ ($\eta_i>0$) and $P_c$ is defined as a standard uniform distribution $\textrm{Unif}(0,1)$. Following~\cite{navon2021learning}, we set $\eta=(0.2,0.2,\dots,0.2)$ for all our experiments. As shown in Table~\ref{table:pref}, varying this parameter on PASCAL-Context (3 tasks) does not have any significant impact on the hypervolume.

\section{Effect of $\tau$}
Setting $\tau \gg \frac{1}{N}$ leads to virtually all tasks being treated as inactive. 
This results in lack of control since there is no active task to be tied to the resource control $c$. Also, there is no compression since the inactive loss searches for active tasks to share nodes with.
Consequently, this mostly leads to a fully branched out network which explains the flat curve in Figure 8. When $\tau = \frac{1}{N}+\epsilon$ ($\epsilon\geq0$), both losses are in effect, which results in proper controllability, but possible ignorance of the uniform (or near uniform) preferences explains the slightly poorer performance. In our experiments, $\tau=0.6/N$ works well across all datasets.

\section{Computational Resource}
\noindent\textbf{Controller overhead.}
The usage of hypernetworks leads to additional resource usage which we term as controller overhead. In comparison to PHN-BN, while we incur a larger overhead due to the prediction of a higher number of normalization parameters, the effect is minimal due to the controller being active infrequently only during preference change. This overhead has no impact on inference.

\noindent\textbf{Model size.}
In this work, we consider inference time as a measure of computational resource. We believe fast inference matters in many practical scenarios, whereas reasonable amount of increase in memory overhead is relatively easy to handle. This justifies our design choice to introduce the anchor net.

\section{Hypervolume}
\begin{figure}[ht]
    \centering
    \includegraphics[width=0.6\columnwidth]{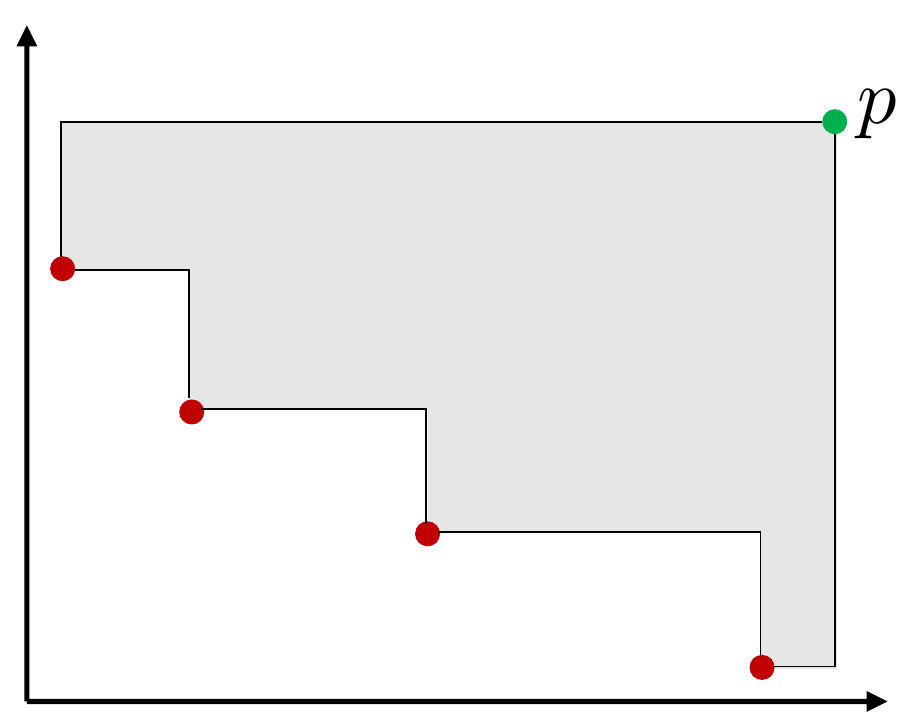}
    \vspace{-0.25cm}
    \caption{\textbf{Hypervolume calculation.} The shaded area represents the hypervolume obtained by the point set comprising the red points, with respect to the reference green point $p$.}
    \label{fig:hv}
    \vspace{-0.3cm}
\end{figure}

Given a point set $S\subset \mathbb{R}^N$ and a reference point $p \in \mathbb{R}^N$ in the loss space, the hypervolume of the set $S$ is defined as the size of the region dominated by $S$ and bounded above by $p$,
\begin{equation}
    \textrm{HV}(S) = \lambda\left(\{a\in\mathbb{R}^N|\exists b\in S:b\preceq a \textrm{ and } b\preceq p\}\right),
\end{equation}
where $\lambda$ is the Lebesgue measure. Figure~\ref{fig:hv} shows an example in two-dimensions.

\section{Potential Negative Societal Impact}
The requirement of heavy computation makes neural architecture search (NAS) environmentally unfriendly. As pointed out in~\cite{strubell2019energy}, the CO2 emissions from a NAS process can be comparable to that from 5 cars’ lifetime. Since our framework is fundamentally a NAS method for multi-task learning, it shares these drawbacks. However, due to the dynamic nature of our method, it saves energy in the long run by allowing a single network to emulate different models corresponding to various preferences.

\end{document}